\documentclass[runningheads]{llncs}

\usepackage{eccv}

\usepackage{eccvabbrv}

\usepackage{graphicx}
\usepackage{float}

\usepackage{amsmath}
\usepackage{amssymb}
\usepackage{booktabs}
\usepackage{multicol,multirow}
\usepackage{setspace}
\usepackage{makecell}
\usepackage[dvipsnames]{xcolor}
\usepackage{tasks}
\usepackage{colortbl}
\usepackage{xcolor}
\usepackage{color}
\usepackage{caption, subcaption, multirow, overpic, textpos}

\usepackage{bbm}
\usepackage{titletoc}

\usepackage{epigraph}
\newlength\savewidth
\newcommand{\tablestyle}[2]{\setlength{\tabcolsep}{#1}\renewcommand{\arraystretch}{#2}\centering\footnotesize}
\renewcommand{\paragraph}[1]{\vspace{1.25mm}\noindent\textbf{#1}}

\newcolumntype{x}[1]{>{\centering\arraybackslash}p{#1pt}}
\newcolumntype{y}[1]{>{\raggedright\arraybackslash}p{#1pt}}
\newcolumntype{z}[1]{>{\raggedleft\arraybackslash}p{#1pt}}

\newcommand{\app}{\raise.17ex\hbox{$\scriptstyle\sim$}}

\definecolor{deemph}{gray}{0.6}

\definecolor{baselinecolor}{gray}{.9}
\newcommand{\baseline}[1]{\cellcolor{baselinecolor}{#1}}
\newcommand{\underfigtab}{\vspace{-10pt}}

\usepackage[dvipsnames]{xcolor}

\definecolor{eccvblue}{rgb}{0.21,0.49,0.74}

\usepackage[pagebackref,breaklinks,colorlinks,citecolor=eccvblue]{hyperref}
\usepackage[capitalize]{cleveref}
\crefname{section}{Sec.}{Secs.}
\Crefname{section}{Section}{Sections}
\Crefname{table}{Table}{Tables}
\crefname{table}{Tab.}{Tabs.}

\usepackage{pifont}% http://ctan.org/pkg/pifont
\usepackage[accsupp]{axessibility}  % Improves PDF readability for those with disabilities.

\usepackage{hyperref}

\usepackage{orcidlink}
\usepackage{tcolorbox}

\title{Embodied Understanding of Driving Scenarios}

\author{Yunsong Zhou\inst{1,2\ast}  \and
Linyan Huang\inst{1\ast} \and
Qingwen Bu\inst{1,2\ast} \and
Jia Zeng\inst{1} \and
Tianyu Li\inst{1} \and
Hang Qiu\inst{3} \and
Hongzi Zhu\inst{2}$^{\dagger}$ \and
Minyi Guo\inst{2} \and
Yu Qiao\inst{1} \and
Hongyang Li\inst{1}$^{\dagger}$
}

\authorrunning{Y.~Zhou et al.}
\institute{
$^1$ OpenDriveLab at Shanghai AI Lab \quad
$^2$ Shanghai Jiao Tong University  \\
$^3$ University of California, Riverside \\ 
[2mm]
\url{https://github.com/OpenDriveLab/ELM}
}

\begin{document}

\maketitle
\begin{center}
    \centering
    \captionsetup{type=figure}
    \includegraphics[width=0.98\textwidth]{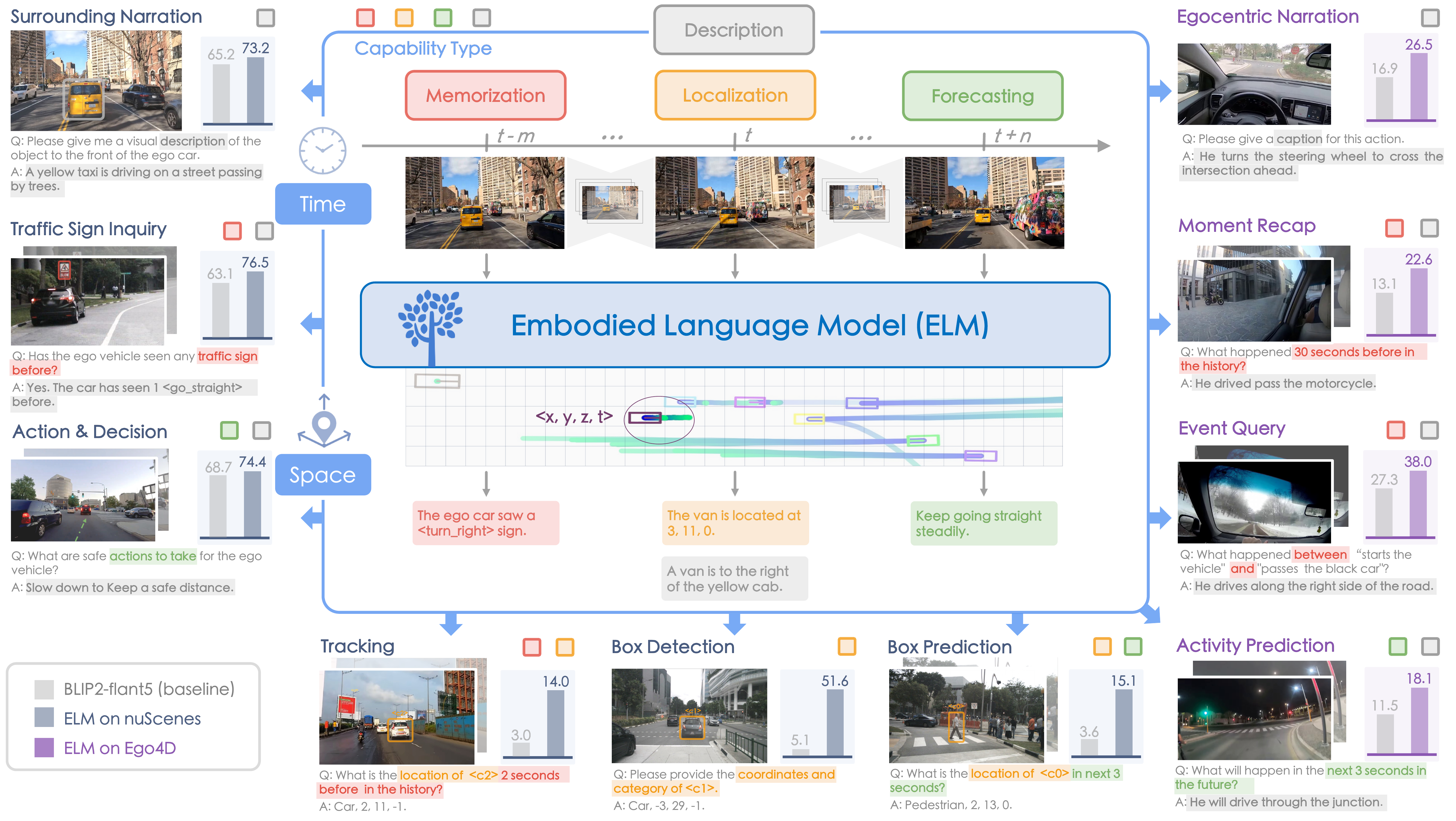}
    \vspace{-3pt}
    \captionof{figure}{
        \textbf{ELM} is an embodied language model for understanding the long-horizon driving scenarios in space and time.
        Compared to the vanilla vision-language model (VLM) being confined to the scene description task, we expand a wide spectrum of new tasks to fully leverage the capability of large language models in an embodiment setting.  
        ELM achieves significant improvements in various applications.
    \label{fig:teaser}
    \underfigtab
    }
\end{center}%

{\let\thefootnote \relax \footnote{$^{*}$Equal contribution. $^\dagger$Co-corresponding authors.}}




\begin{abstract}
Embodied scene understanding serves as the cornerstone for autonomous agents to perceive, interpret, and respond to open driving scenarios.
Such understanding is typically founded upon Vision-Language Models (VLMs). 
Nevertheless, existing VLMs are restricted to the 2D domain, devoid of spatial awareness and long-horizon extrapolation proficiencies.
We revisit the key aspects of autonomous driving and formulate appropriate rubrics.
Hereby, we introduce the Embodied Language Model (ELM), a comprehensive framework tailored for agents' understanding of driving scenes with large spatial and temporal spans.
ELM incorporates space-aware pre-training to endow the agent with robust spatial localization capabilities.
Besides, the model employs time-aware token selection to accurately inquire about temporal cues.
We instantiate ELM on the reformulated multi-faced benchmark, and it surpasses previous state-of-the-art approaches in all aspects. 
All code, data, and models 
% will be publicly shared.
are accessible.
\end{abstract}

\section{Introduction}
\label{sec:intro}

% === Epigraph ===
% \epigraph{
%     \textit{Time and space are modes by which we think and not conditions in which we live.}
% }{
%     \textsc{--- Albert Einstein}
% }
% \begin{center}\rule{0.7\linewidth}{0.3pt}\end{center}

% \vspace*{1mm}

% \input{sec/0_abstract}    
% \input{sec/1_intro}
% \input{sec/2_related_work}
% \input{sec/3_finalcopy}
% \input{sec/4_conclusion}

Embodied understanding enables intelligent agents (\textit{e.g.}, self-driving vehicles, robots, and drones) to interpret instructions and analyze scenes based on their experience~\cite{grauman2022ego4d,yang2023survey}.
However, this critical but challenging task is yet to be solved.
Recently, benefiting from their extensive knowledge and causal reasoning capability, vision language models (VLMs)~\cite{li2023blip2,zhang2023llamaadapter,alayrac2022flamingo,liu2023llava} have achieved remarkable progress in general vision~\cite{lu2022learn,chen2023tem,2023videochat,li2023mimicit,Maaz2023VideoChatGPT}.
The utilization of VLMs provides a question-answering framework to engage with a scene and contribute to common sense comprehension.
When it comes to driving scenarios, embodied approaches via VLMs have the potential to surpass both rule-based~\cite{voigt2017eu,sauer2018conditional,fan2018baidu,regulation2020art} and data-driven learning-based~\cite{casas2021mp3,hu2022stp3,hu2023_uniad,dauner2023parting} methods in unforeseen scenarios \cite{liu2023llava,chen2023drivingwithllms,yang2023survey}.

% As an inherently embodied agent, the autonomous driving (AD) vehicle urgently requires an explicable and all-round comprehension of the visual world for informed planning. 

To cope with complex driving scenarios, it is crucial for an embodied agent to obtain a complete 4D scene understanding, particularly in extensive spatial scale and extended temporal duration.
As depicted in \cref{fig:teaser}, this calls for four pivotal capabilities, including \textbf{1)} \emph{description}: the agent is able to describe the surrounding environments; \textbf{2)} \emph{localization}: rather than merely assessing approximate position, the agent needs to pinpoint a particular object in the 3D space; \textbf{3)} \emph{memorization}: the agent needs to retrieve specific events that have occurred; and \textbf{4)} \emph{forecasting}: the agent is required to foresee a certain future from the given history.

%  Contemporary efforts exploit VLMs for numerous scene understanding tasks in general vision, including visual interpretation \cite{lu2022learn,chen2023tem,2023videochat,li2023mimicit,Maaz2023VideoChatGPT}, open-vocabulary visual recognition \cite{chen2021pix2seq,kirillov2023segment,minderer2205simple,radford2021learning}, object-agent interactions \cite{chen2023pali,hao2022language,huang2023language,wang2022git}, and \textit{etc}.
%
% In driving scenarios, pioneering work is inspired to propose the use of VLMs to achieve a descriptive understanding of the content of the scene through the question-answer format.

Recently, attempts are conducted to incorporate VLMs into the autonomous driving domain.
Current methods are instrumental in crafting narrations encompassing the surroundings environment~\cite{malla2023drama}, traffic participants~\cite{deruyttere2022talk2car}, road components~\cite{Dewangan2023talk2bev}, potential interactions~\cite{kim2019CVPR,wu2023nuprompt}, and driving behaviors~\cite{kim2018textual,sachdeva2023rank2tell,xu2020explainable}. 
Nevertheless, the capabilities of vanilla VLMs are limited to generating narrative phrases, namely description.
% The ability of VLMs to sense space and time remains unexplored, as existing works can only describe approximate positions \cite{qian2023nuscenes} and achieve information retrieval in a short
% period \cite{li2023mimicit,2023videochat}. 
%
% But, the capabilities of vanilla VLMs are limited to generating narrative phrases, \textit{i.e.} description.
% We refer to such narrative capabilities collectively as description.
Their sense of space and time remain unexplored, as existing works can only describe rough position information \cite{qian2023nuscenes} and achieve information retrieval in a short period \cite{li2023mimicit,2023videochat}. 
% Lack of relevant data and necessary modules, these approaches exhibit deficiencies in spatial perception \cite{qian2023nuscenes}, and they can only achieve information retrieval in short
% time scales.
%
% Their understanding of the world is incomplete, and thus, they do not cope with the task of embodied understanding for driving scenarios.
%
As such, the absence of localization, memorization, and forecasting refrains VLMs from the embodied understanding of driving scenarios.
% 
% \cite{yang2023survey}.

% To address the newly expanded task of embodied understanding,
% In this paper, 
To this end, we introduce \textbf{Embodied Language Model (ELM)} for the proposed driving scene understanding problem.
% with large spatial and temporal spans. 
% empirically outperforming 
% beyond
% prior approaches for driving scene understanding.
%
As highlighted in \cref{fig:teaser}, in contrast to conventional VLMs which only possess description capability, ELM extends the capabilities of language models in large spatial and temporal horizons.
%
% Along with the freshly 
Amongst the newly formulated problem, a suite of tasks and evaluation protocols are presented.
% to assess the level of scene understanding.
%
The key challenges are presented as follows:

\smallskip
\noindent\textbf{Long horizon in Space.}
Since VLM decoders are natrually insensitive to numbers, an intuitive solution would be to rewrite the vocabulary~\cite{rt22023arxiv} and pre-train models on numerically relevant tasks with the replaced words.
%
%Contemporary approaches \cite{driess2023palme,rt22023arxiv} provide good insights and achieve impressive performance.
%
However, excessive training on a single type of data may lead to catastrophic forgetting \cite{zhai2023investigating}.
Data diversity is therefore of crucial importance.
%
%
%
% To tackle this challenge, w
We propose \textit{space-aware pre-training} along with a diverse data collection and auto-labeling process.
We orchestrate over 3,000 hours of data and 9 million pairs of diverse annotations from the open world, incorporating the public autonomous driving datasets nuScenes \cite{caesar2019nuscenes} and Waymo \cite{sun2020scalability}, the internet-derived dataset YouTube and the egocentric dataset Ego4D \cite{grauman2022ego4d}.
This enables the autonomous agent to acquire spatial localization competence while preserving the initially robust descriptive aptitudes.

\smallskip
\noindent \textbf{Long horizon in Time.}
Summarizing long historical time-series data is computationally burdensome with significant redundancy.
%
% Given the resource constraint on the deployment of VLMs, 
A straightforward way is to split and sample the video into images~\cite{2023videochat,li2023mimicit,Maaz2023VideoChatGPT}.
While there is an attempt to summarize a film as a sequence of chronologically occurring events~\cite{song2023moviechat}, it does not allow the agent to recall events from a brief moment in a lengthy video.
%
% While some work \cite{song2023moviechat} has attempted to summarise a film using sparse tokens, the agent only learns to produce generalized descriptions in the order of events and cannot pinpoint what happened at any particular second in a long memory.
%
We are of the opinion that the crux lies in enabling the agent to efficiently retrieve the most pertinent content from long-term memory based on the given instruction.
To accomplish this, we opt in a module named \textit{time-aware token selection}.
%
%a module designed to allow the agent to have a notion of the event timestep and selectively fetch the tokens used.
%
The module encodes each frame into sparse tokens and builds a token bank.  
A set of learnable queries is leveraged to extract the most relevant moment-specific and content-specific cues emphasized in the instruction,  enabling effective long-term information retrieval.

\smallskip
\noindent \textbf{Benchmark.}
%To put the aforementioned competencies into action, 
To evaluate ELM and other VLMs, we assemble a new evaluation suite comprising ten distinct tasks. 
These tasks encompass evaluations of both individual and integrated competencies in description, localization, memorization, and forecasting, as delineated in \cref{tab:tasks}. 
%
% These tasks cover the assessment of individual competencies and combinations of competencies, as shown in Tab. \cref{tab:tasks}. 
% 
% We revisit the specific tasks needed to assess each competency of autonomous driving scenarios, and we list the recommended ten tasks in Tab. \cref{tab:tasks}.
%
%Some added tasks complement the limitations of vanilla VLMs in the understanding of driving scenarios.
%
%As illustrated in Fig. \cref{fig:teaser}, these tasks include both the vanilla VLM's ability to describe the scene (top left), while extending the model's localization for space (bottom), and memory for time information (right).
%
The devised tasks include descriptive tasks within the purview of vanilla VLMs, as well as tasks involving spatio-temporal localization and dynamic information prediction. While the primary focus of this investigation pertains to driving scenarios, it is worth noting that the incorporation of daily indoor scenarios can serve as a valuable means to assess VLMs' capacity for long-term event reasoning.
The details of the formulated tasks are described in \cref{sec:preliminary}.

% %
% ELM is designed in the spirit of embodiment, encapsulating open tasks in the form of natural languages.
% %
% We are not simply transposing existing VLMs to driving scenarios, but rather focusing on addressing the innate disabilities that impede the application of VLMs to autonomous driving.
% %
% The key component of our ELM lies in the spatio-temporal tokenizer.
% %
% It forms a scene description accompanied by location information from both space and time attributes, enabling language models to recognize events occurring in four-dimensional scenarios.
% %
% Moreover, the sparsified token representation and inquiry-driven approach allow VLMs to achieve long-range scene understanding with less effort.

The \textbf{contributions} are three folds:
\textbf{a)} We revive driving scene understanding by delving into the embodiment philosophy.
This involves a deconstruction of its definition and basic capabilities, along with a series of novel tasks and a comprehensive evaluation benchmark.
\textbf{b)} We propose ELM, a vision-language model for embodied understanding in driving scenarios.
% \textbf{(a)} we revive autonomous scene comprehension by ELM, a vision-language model for embodied understanding in driving scenarios.
%
Our proposed space-aware pre-training strategy and time-aware token selection enhance agents' comprehension in long-range four-dimensional space.
%
% \textbf{(b)} we build up a systematic evaluation benchmark by involving a series of novel tasks in scene understanding.
%
%
% Our proposed space-aware pre-training strategy and time-aware token selection enhance agents' comprehension in long-range four-dimensional space.
% \textbf{(a)} we revive autonomous scene comprehension by adhering to the embodiment philosophy. This involves creating a  series of novel tasks and a comprehensive evaluation benchmark.
% %
% \textbf{(b)} we propose ELM, a vision-language model for embodied understanding in driving scenarios.
% %
% Our proposed space-aware pre-training strategy and time-aware token selection enhance agents' comprehension in long-range four-dimensional space.
%
% data
\textbf{c)} We validate ELM on the all-encompassing tasks for cross-domain scenarios. 
Experimental results demonstrate the superiority of our method compared to LLaMA-Adapter V2 \cite{gao2023llamaadapterv2}, LLaVA \cite{liu2023llava}, Otter \cite{li2023mimicit}, VideoChat \cite{2023videochat}, \textit{etc}.
\cref{fig:teaser} visualizes the achieved improvement compared to BLIP2-flant5~\cite{li2023blip2} across ten distinct tasks.

\begin{table}[t!]
\centering
%\footnotesize
%\tablestyle{2.0pt}{1.05}
% \vspace{-.2cm}
% \setlength{\tabcolsep}{2mm}{
\scalebox{0.63}{
\begin{tabular}{l|c|cccc|ccc}
\toprule
\multirow{2}{*}{Tasks} & \multirow{2}{*}{Fine-tune Dataset} & \multicolumn{4}{c|}{Capability}              & \multicolumn{3}{c}{Statistics}  \\
                     &  & Description  & Localization & Memorization & Forecasting & S(\textit{m}) / R(\textit{m}) & T (\textit{s}) / F & \# \\
\midrule 
\color{gray}{Surrounding Narration}   &    \multirow{6}{*}{nuScenes \cite{caesar2019nuscenes}}      &          \color{gray}{\checkmark}  &          &             &           & \color{gray}{30 / 5}              & \color{gray}{0.5 / 1}    & \color{gray}{142K}        \\
Traffic Sign Inquiry      &       &    \checkmark       &          & \checkmark           &           & 30 / 1           & 3.5 / 7    & 20K        \\
Action \& Decision   &    &  \checkmark     &          &             & \checkmark         & 30 / 5              & 3.5 / 7  & 301K       \\
Box Detection         &         &       &       \checkmark       &             &           & 50 / 1           & 0.5 / 1    &  232K        \\
Tracking    &       &         &     \checkmark        & \checkmark           &           & 50 / 1           & 3.5 / 7    & 131K       \\
Box Prediction       &         &        &    \checkmark         &             & \checkmark         & 50 / 1           & 3.5 / 7    & 133K        \\

\midrule
\color{gray}{Egocentric Narration}       &  \multirow{4}{*}{Ego4D \cite{grauman2022ego4d}}     &   \color{gray}{\checkmark}       &           &             &           & \color{gray}{20 / 3}              & \color{gray}{3 / 1}      &  \color{gray}{357K}     \\
Moment Recap   &        &    \checkmark       &          & \checkmark           &           & 20 / 3              & 60 / 20   & 70K       \\
Event Query       &     &     \checkmark     &           & \checkmark           &           & 20 / 3              & 60 / 20    & 70K      \\
Activity Prediction    &     &   \checkmark  &           &             & \checkmark         & 20 / 3              & 60 / 20  & 69K   \\ 
\bottomrule
\end{tabular}}
\vspace{5pt}
\caption{
\textbf{Performing Tasks for Embodied Understanding of Driving Scenarios. }
% We build up an evaluation system comprising the ten tasks listed above. 
We supplement the evaluation of long-term memory with long videos from Ego4D \cite{grauman2022ego4d}, which is lacking in self-driving datasets.
The \color{gray}{gray-colored tasks} \color{black}{are} already applicable to vanilla VLMs. 
\texttt{S}: the span in space; \texttt{R}: the resolution in space; \texttt{T}: total duration; \texttt{F}: the number of frames; \texttt{\#}: the number of QA pairs.
}
\label{tab:tasks}
\underfigtab
\end{table}

\section{Problem Setup}
\label{sec:preliminary}

%An extensive collection of tasks is needed to facilitate and evaluate embodied understanding in driving. 
%
%\noindent \textbf{Key highlight of the task.}
%
%Within the realm of embodied understanding, textual question-answering (QA) plays a human-centric approach between drivers and autonomous agents, wherein intricate queries are simplified into interpretable outputs. 
%
%Through the utilization of such agents, one can attain natural interactions, thereby bolstering transparency and trust in autonomous driving vehicles and consequently enhancing their explainability and accountability.
%
Based on the analysis of the pivotal competencies involved in embodied understanding, the newly proposed benchmark thoroughly evaluates VLMs from the perspective of description, localization, memorization, and forecasting.
Utilizing the nuScenes~\cite{caesar2019nuscenes} and Ego4D~\cite{grauman2022ego4d} datasets, we formulate ten question-answering (QA) tasks as listed in \cref{tab:tasks}.
%to evaluate the aforementioned four capabilities. The designations and statistics of these tasks are listed in \cref{tab:tasks}.
% 

Built on top of the nuScenes dataset, we present three tasks which are for prompting embodied agents to provide descriptions of the current scene, recall previously observed traffic elements, and predict future states.
Furthermore, we devise three localization-related tasks, which require embodied agents to deduce the 3D positions of 2D query points in the present, past, and future. Completing these positioning tasks necessitates robust spatial perception and temporal reasoning capabilities.
%
%Additionally, we formulate three object location tasks,  detection QA tasks for accurate spatial localization within +/- 50 meters forefront view with 1-meter resolution, encompassing classification as well. 
%
% To ensure that VLMs remain unbiased towards autonomous driving scenes, the proposed benchmark incorporates a diverse range of data from both common 
% %~(\textit{i.e.}, Ego4D dataset\cite{grauman2022ego4d}) 
% and driving-specific scenarios. 
To ensure that VLMs remain unbiased towards driving scenes, we incorporate the Ego4D dataset for evaluation in common scenarios. 
% In addition to evaluating on driving dataset, 
% We carefully curate tasks using the Ego4D dataset \cite{grauman2022ego4d}. 
These tasks require the description of ongoing events, inquiry of past events, and prediction of future events.
The scenes in Ego4D consist of prolonged videos, and this necessitates a greater understanding over long time spans.

%
%complementary to nuScenes \cite{caesar2019nuscenes}. 
%
%The Egocentric Narration task entails the description of scenes depicted in single-frame visuals.
%
%The Moment Recap, Event Query, and Activity Prediction tasks, on the other hand, leverage up to 60 seconds of video input. 
%
%These tasks involve pinpointing specific past video events via timestamps, addressing blanks between consecutive events, and predicting future occurrences.

The formulation of each task is elaborated as follows:

% \begin{enumerate}
%     \item 1
%     \item 1
% \end{enumerate}

\begin{itemize}
    % \item \textit{Surrounding Narration:} provides an overall description of the surroundings. Specifically, this task aims to analyze the presence, movement, and occlusion of traffic objects using a single frame.
    \item \textit{Surrounding Narration:} providing an overall description of the surroundings, namely attribute, presence, and movement of traffic objects on a single frame. 
    %
    % \item \textit{Traffic Sign Inquiry:} recalls traffic signs observed previously. With a 3.5-second video input, the model is required to provide information on past traffic signs and lane markings.
    \item \textit{Traffic Sign Inquiry:} identifying and recalling traffic signs and lane markings observed within 3.5 seconds in the past.
    %
    % \item \textit{Action \& Decision:} provides a high-level planning-related instruction text. Given a 3.5-second video, the model needs to anticipate potential interactions and make driving decisions.
    \item \textit{Action \& Decision:} providing a high-level planning-related instruction to foresee potential interactions and make driving decisions.
    %
    % \item \textit{Box Detection:} infers the 3D location and category of a 2D-pixel query. The model tasks a single frame as input and is expected to provide the 3D coordinate and category based on the 2D query point.
    \item \textit{Box Detection:} inferring the 3D coordinate and category based on the 2D query point on a single frame.
    %
    % \item \textit{Tracking:} retrieves the previous 3D location of a 2D-pixel query.  Given a 3.5-second video, a past timestamp, and the current 2D position of the queried object, the model infers its 3D coordinate and category.
    \item \textit{Tracking:} retrieving the 3D trajectory and category of the object queried by the 2D pixel position of the current frame for the last 3.5 seconds.
    %
    % \item \textit{Box Prediction:} infers the future 3D location of a 2D-pixel query. Provided with the identical input format as in \textit{Tracking} task, the model outputs the 3D coordinate at a future moment and the object's category.
    \item \textit{Box Prediction:} inferring the future 3D location and category of a queried object given its current 2D coordinate and 3.5s past observations.
    \item \textit{Egocentric Narration:} describing self-behaviors~(actions and interactions with the surroundings) based on an egocentric single-frame input. 
    %
    % \item \textit{Moment Recap:} pinpoints an event that occurred at a past moment. The model is required to deduce the event that occurred at a past moment within a 60-second video.
    \item \textit{Moment Recap:} indicating an event that occurred at a specific point in time within the last 60 seconds.
    %
    % \item \textit{Event Query:} denotes an event that occurs between two specified events. Utilizing a 60-second video as input, the model infers the content of a specific event through the analytical examination of two contiguous events: one antecedent and one subsequent to the target event.
    \item \textit{Event Query:} deducing the content of an specific event through the analytical examination of its antecedent and subsequent events in a 60-second video.
    %
    % \item \textit{Activity Prediction:} predicts an event that will happen in a future time. Through analyzing a 60-second video, the model is tasked with forecasting the event that will take place at a designated future moment.
    \item \textit{Activity Prediction:} predicting an event that will happen at a designated future moment in a 60-second video.
\end{itemize}
%
% For the Box Detection task, a single frame serves as the input, wherein the model is expected to provide the 3D location and category based on the 2D-pixel query. 
%
% Tracking, on the other hand, involves retrieving the previous objects' 3D locations from their current 2D positions, while Box Prediction focuses on predicting the future locations of 3D objects. 
%
% For instance, a sample inquiry could be, ``Q: what was the location of $\langle c0, 487, 231 \rangle$ 3 seconds ago? A: Location: [-3, 35, 0], Category: Pedestrian.''
%
%Furthermore, the \textit{Surrounding Narration} task aims to analyze the presence, movement, and occlusion of traffic objects using a single frame.
%

%For \textit{Traffic Sign Inquiry} and \textit{Action \& Decision} tasks,  Additionally, the model is designed to formulate reasoned courses of action grounded in potential circumstances.
%
\begin{figure*}[t]
  \centering
  \includegraphics[width=\linewidth]{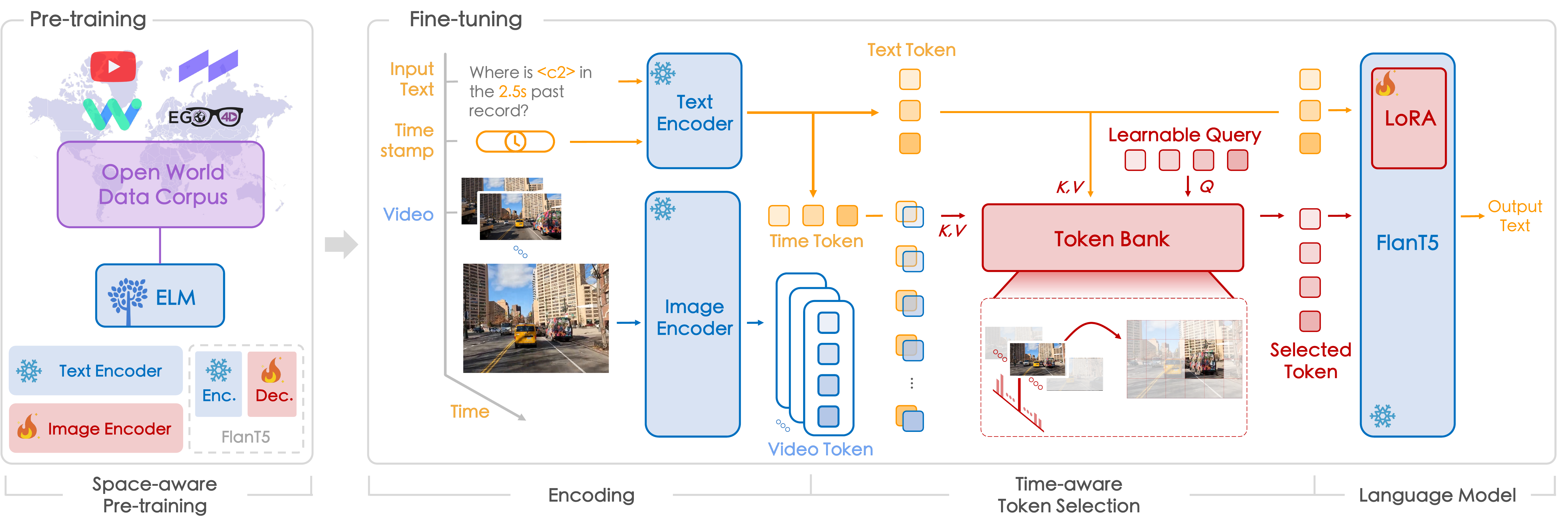}
  \vspace{-15pt}
   \caption{
   % The overview of our methods. 
   \textbf{Systematic Pipeline of ELM.}
   % Our pipeline 
   It consists of Pre-training by open-world data corpus and Fine-tuning on diverse tasks.
   To initialize the Space-aware Pre-training, we collect extensive image-text pairs from the world, empowering ELM with spatial localization while preserving the description ability in driving scenarios.
   In the fine-tuning process, the inputs to ELM are videos, timestamps, and text prompts.
   After encoding the inputs into tokens, ELM leverages the proposed Time-aware Token Selection to gather the appropriate tokens as instructed by prompts.
   Finally, the tokens are sent to the language model to generate output texts.
   % , and we employ LoRA \cite{hu2021lora} for expedited adaptation.
   }
   \label{fig:overview}
   \underfigtab
\end{figure*}

% The 
% % above 
% tasks above are summarized as our newly proposed criterion for the embodied understanding of driving scenes.
% % Presenting a collection of 
% % Equipped with the collection of these tasks, we introduce an evaluation benchmark.
% %that encompasses both spatial and temporal proficiencies. 
% %
In contrast to previous datasets and tasks~\cite{qian2023nuscenes,wu2023nuprompt,sachdeva2023rank2tell,malla2023drama,kim2018textual,kim2019CVPR,sima2023drivelm}, the proposed benchmark incorporates both spatial and temporal evaluation, necessitating embodied agents to have a correct understanding of the complex driving scenes.
We set up this benchmark for assessing embodied understanding in driving scenarios and harmonizing diverse driving-related objectives.

\smallskip
\noindent \textbf{License and privacy considerations.} 
\label{sec:license}
All the annotated data (benchmarks and open-world data corpus mentioned in \cref{sec:space_aware_pre_training}) comply with the CC BY-NC-SA license.
Following \cite{abu2016youtube,xu2018youtube,zhang2022learning,zhu2022celebvhq,kay2017kinetics}, we safeguard the rights of the data owners and prevent privacy leakage by distributing redirection links instead of publishing image contents. 
Personal identification information will not be leaked to the public.
More details are shown in the Appendix.

\section{Methodology}

\subsection{Overview}
We aim at enhancing agents' spatial perception with diverse pre-training data and dealing with long time series through adaptive token selection.
% The 
% % guiding 
% philosophy behind ELM is to utilize diverse data  during 
% pre-training to enhance agents' spatial perception and employ adaptive selection for tokens to deal with long-time series.
%
\cref{fig:overview} illustrates the architecture of our framework.
%
% The overall training process consists of two stages, pre-training, and finetuning, each designed to incrementally develop description, localization, and memorization/forecasting capabilities.
%
ELM begins with Space-aware Pre-training (\cref{sec:space_aware_pre_training}) on image-text pairs. 
During this phase, ELM focuses on the vocabulary related to spatial understanding and learns a robust visual encoder through extensive training.
Throughout the fine-tuning across varied tasks, all encoders of ELM are frozen.
%
% After freezing all encoders, ELM receives input video, timestamp, and text prompt during the Fine-tuning on diverse task.
%
In the Encoding process, the text prompt and timestamp are encoded by BERT~\cite{devlin2018bert}, while each video frame is transformed into fixed-length feature tokens using the EVA model~\cite{fang2023eva}.
In Time-aware Token Selection (\cref{sec:time_aware_token_selection}), the video tokens are fed into a token bank along with the text and timestamp tokens, and the token bank adaptively selects the desired tokens based on the text prompt.
%
% The selected tokens are of fixed length, regardless of the video's duration.
%
% {\color{red} LoRA is more about a training technique.}
Lastly, the FlanT5~\cite{2020t5} model, fine-tuned with LoRA~\cite{hu2021lora}, generates the output text to tackle various tasks in our benchmark.

\begin{table}[t]
\centering
% \footnotesize
\tablestyle{3.0pt}{1.05}
\scalebox{0.73}{\setlength{\tabcolsep}{2mm}{
\begin{tabular}{l|c|cc|cc|c}
\toprule
\multirow{2}{*}{Method} & \multirow{2}{*}{Pre-train Data} & \multirow{2}{*}{\# Frames} & Duration  & Geographic  & Diversity &  \multirow{2}{*}{Anno} \\
 & & & (hours) & Countries & Cities & \\
\midrule 
LLaVA \cite{liu2023llava} & COCO \cite{lin2014coco} & 150K  & - &- & - & Des \\ 
VideoChat \cite{2023videochat} & Self-Collected & 18K & - & - & - & Des \\ 
Vid-ChatGPT \cite{Maaz2023VideoChatGPT} & ActivityNet-200 \cite{caba2015activitynet} & 100K & - & - & - & Des \\ \midrule
nuScences-QA \cite{qian2023nuscenes} & nuScenes \cite{caesar2019nuscenes}  & 460K & 5.5 & 2 & 2 & Des \\
DriveGPT4 \cite{xu2023drivegpt4} & BDD-X \cite{kim2018textual} & 28K & 77 & 1 & 4 & Des \\
LLM-driver \cite{chen2023drivingwithllms} & Self-Collected & 160K & - & - & - & Des 
\\ 
DriveLM \cite{sima2023drivelm} & nuScenes \cite{caesar2019nuscenes}, CARLA \cite{Dosovitskiy17} & 188K & 95 & 3 & 3 & Des \\
% \multirow{2}{*}{DriveLM \cite{sima2023drivelm}} & nuScenes \cite{caesar2019nuscenes} & 5K & 5.5 & 2 & 2 & Des \\
%  & CARLA \cite{Dosovitskiy17} & 183K & 90 & 1 & 1 & Des \\
\midrule
\multirow{4}{*}{\textbf{ELM (Ours)}} & nuScenes \cite{caesar2019nuscenes}    &   7.4M   &   5.5      &  2 & 2      & Des, Loc     \\
& Waymo \cite{sun2020scalability}     &    450K   &    6.4      &  1  &  6      & Des         \\
& YouTube      &    1.1M     & 1474       &  $\geq$40 &    $\geq$ 709      &     Des     \\
& Ego4D \cite{grauman2022ego4d}   &    300K      &     1638    & 9 & 74          &    Des     \\
\bottomrule
\end{tabular}
}}
\vspace{5pt}
\caption{
\textbf{Statistics of pre-training data and comparison of data collection with other VLMs.} 
Our pre-train data surpasses that in general vision (top) and autonomous driving (middle) in terms of quantity and diversity. \texttt{Anno}: the type of annotations; \texttt{Des}: description; \texttt{Loc}: localization.
}
\vspace{-20pt}
\label{tab:pre-train}
% \underfigtab
\end{table}

% \smallskip

\subsection{Space-aware Pre-training}
\label{sec:space_aware_pre_training}

\noindent \textbf{Open world data collection.}
In pursuit of spatial localization while retaining the description capacity for driving scenarios, we collect an open-world data corpus for the space-aware pre-training.
As depicted in \cref{tab:pre-train}, the data corpus is derived from a variety of sectors.
Representative datasets for autonomous driving, such as nuScenes \cite{caesar2019nuscenes} and Waymo \cite{sun2020scalability}, constitute our fundamental resources.
%
% Collectively, 
These two datasets comprise a total of 11.9 hours of data, capturing scenes from five different cities: Boston, Singapore, San Francisco, Phoenix, and Mountain View.
YouTube, renowned for its extensive data and diverse content, serves as a critical resource for our research.
%
% Via 
We collect a total of 1,474 hours of publicly available videos from over 709 cities in more than 40 countries using web crawlers. 
The collected data covers a wide range of locations, including urban areas, rural regions, and various weather conditions.
% 
% varied locales like urban areas, fields, \textit{etc.}, and different weather conditions.
%
For a broader vision, we utilize the Ego4D dataset \cite{grauman2022ego4d}, which provides an in-depth understanding of daily activities worldwide.
There are 931 camera wearers contributing a total of 1,638 hours of footage from 74 cities. 
%
% To comprehend autonomous driving, 
We aggregate an extensive and diverse dataset for pre-training, which goes far beyond those adopted in other VLMs~\cite{liu2023llava,2023videochat,Maaz2023VideoChatGPT,qian2023nuscenes,xu2023drivegpt4,chen2023drivingwithllms}.
% \cite{liu2023llava}

\begin{figure}[t]
    \centering
    \includegraphics[width=0.99\linewidth]{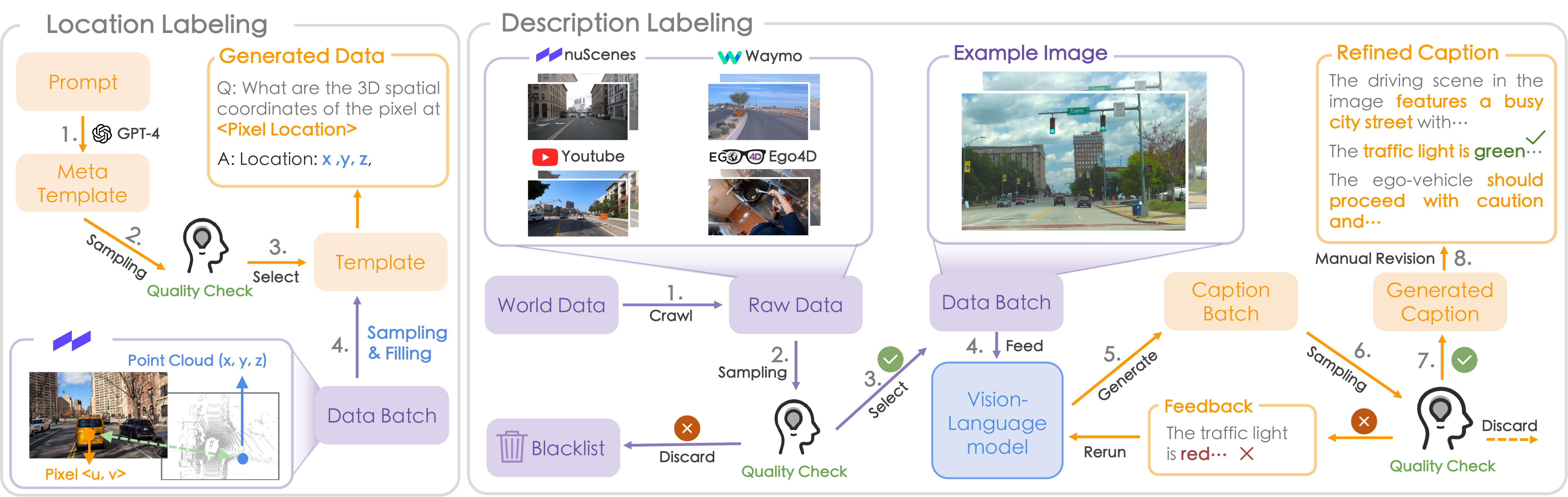}
    \vspace{-5pt}
    \caption{\textbf{
    % The data collection pipeline.
    Annotation workflow
    % Description labeling pipeline 
    with human quality check in the loop.
    } 
    \textbf{For location labeling: }we first select diverse templates from the GPT generated candidates. Pixel-point pairs as annotated in the nuScenes~\cite{caesar2019nuscenes} are then sampled and filled into the templates to form our location pre-training data.
    % We take the description labelling task as an example. 
    \textbf{For description labeling: }
    \texttt{Node} 4 utilizes LLaMA-Adapter V2 \cite{zhang2023llamaadapter} to obtain diverse labels on nuScenes, Waymo \cite{sun2020scalability}, YouTube, and Ego4D \cite{grauman2022ego4d} with predefined prompts.
    % A dual 
    Two rounds of 
    quality check are conducted in 
    \texttt{Node} 3 and 7
    by inspectors to guarantee the image and caption quality.
    % production of images and labels 
    % with exceptional quality.
}
    \label{fig:data_gen}
    \underfigtab
\end{figure}

\smallskip
\noindent 
% \textbf{Human-assisted auto-labeling.}
\textbf{Auto-labeling with human in the loop.}
% As depicted in the upper part of \cref{fig:data_gen},
% To facilitate the understanding of space,
% for the autonomous agent, 
With the objective of enhancing models' spatial comprehension,
we design a localization labeling process based on nuScenes \cite{caesar2019nuscenes} in \cref{fig:data_gen}.
%
% Considering the superior performance of GPT-4 compared to crowdsourcing workers in text annotation \cite{gilardi2023chatgpt}, 
To ensure the diversity of questions, 
we use GPT-4 \cite{openai2023gpt} to generate massive unique templates for text prompts.
% to ensure diversity.
%
In response to the GPT's instability, we execute a manual selection process to assemble a set of 1000 high-quality templates.
Regarding the location ground truth labels, we leverage the point clouds and camera parameters to establish the correspondence between 2D pixels and 3D point coordinates. 
In addition, we employ density-based point sampling to achieve uniform coverage in 3D space, followed by a rule-based method to assign the labels to the templates.
Collectively, we create a total of 7.4M QA pairs about location. The pipeline is detailed in the Appendix.

For preventing catastrophic forgetting during pre-training \cite{zhai2023investigating}, we introduced a large number of description labels into the data corpus, thereby enhancing the diversity of the dataset.
% we augment the diversity of the data corpus through the inclusion of an extensive array of descriptive labels.
%
The right side of \cref{fig:data_gen} illustrates our description labeling pipeline, and the labels include descriptive sentences of the overall scenario, transport elements, and driving decisions.
Particularly, two rounds of quality check are implemented to maintain a high standard of labeled data.
The annotation pipeline starts by removing noisy, interfering, and blurry images from \texttt{Node} 1 to \texttt{Node} 3.
After crawling raw data from the open world, the inspectors extract 10\% of a batch of images for a quality check to determine if the batch should be retained.
The image selection process primarily involves sorting out the worst $N$ samples in terms of quality from a quantitative set of video clips based on standards like lighting, resolution, and clarity. 
These are then returned to the reserve pool, with the remainder forwarded to the next process. 
If a video source is repeatedly flagged as poor quality, it is placed on a blacklist.
%
% After selecting the data batch, 
The qualified data batches are fed into LLaMA-Adapter V2 \cite{gao2023llamaadapterv2} to generate caption batches, while others are discarded.
% the second quality check is utilized on the caption generated by.
Following this, the second quality check on the generated caption is performed in \texttt{Node} 6-7. The revised captions will be saved as the final annotations in \texttt{Node} 8.
In instances where a data batch fails to meet quality standards, inspectors will furnish feedback to the model for the purpose of refining the generated captions in subsequent iterations.
Labeling details, discarded images, and annotation examples are in the Appendix.

% Starting by removing noisy, interfering, and blurry images, the pipeline .
% %
% For each image batch collected from the Internet, the inspectors extract 10\% of it for quality check.

% A quality check loop with inspectors is then conducted to ensure the high standard of the caption data generated by LLaMA-Adapter V2 \cite{gao2023llamaadapterv2}.
%
% Inspectors first randomly sample 10\% of the generated caption within a data batch. 
% %
% These captions will they be saved as final annotations only when they meet the requirements, otherwise this batch of data will be re-sent to the model for adjustment and refinement.

% As such, 
Following the workflow above, we have amassed over 9 million annotations, 
% which reflect 
indicating a substantial increase in the scale and diversity of the dataset used for pre-training.
% {\color{red}
% The comparison with prior art \cite{qian2023nuscenes,wu2023nuprompt,sachdeva2023rank2tell,malla2023drama,kim2018textual,kim2019CVPR} training on limited data are illustrated in \cref{sec:diversity_data}. remove}
%
The comparison of annotation quality and diversity will be further demonstrated in~\cref{sec:discussion} and the Appendix.

\smallskip
\noindent \textbf{Tokenizer.} 
\label{sec:tokenizer}
%
% Given 
It is argued
that VLMs are insensitive to numbers~\cite{driess2023palme}. An RT2-like tokenizer \cite{rt22023arxiv} is implemented to enable a general VLM to perform location prediction in the form of text.
% :
% %
% \begin{equation}
%     \mathcal{L} = \mathcal{K}([x,y,z, c]) = [w_1, ... w_n],
% \end{equation}
% %
% where $x$, $y$, $z$ are coordinate value in 3D space, $c$ is the object category, $w_i$ is the $i$-th word in this sentence.
% % 
% Note that each coordinate value can be freely transformed into a set of words using the tokenizer $\mathcal{K}$.
%
We divide the 3D space into 1-meter resolution grids and quantify the position of the target point as the index of the grid. 
Then we rewrite the least frequently used words in FlanT5 
% \cite{2020t5} 
to represent the grid index, which is referred to as space-relevant vocabulary.
%
% Consequently, 
Hence 
% we could reformulate 
the 3D localization 
% problem 
could be deemed as a language modeling task.

\subsection{Time-aware Token Selection}
\label{sec:time_aware_token_selection}

To effectively memorize and forecast events in long time-series videos, it is essential to 
encode the scene using 
a timestamp-sensitive representation and select tokens wisely. 
%
% As shown in~\cref{fig:module}, 
Thus, we introduce the Time-aware Token Selection module, which utilizes the input text prompt as guidance to select a fixed number of relevant tokens from the video. 
These selected tokens are then incorporated into the language model as visual input. 
To facilitate interaction among videos, timestamps, and prompts, it is important to align their embeddings within the textual feature space, for which we perform the following design.

\smallskip
\noindent \textbf{Video encoding.}
Initially, we utilize Q-former \cite{li2023blip2} to align the video features with language model inputs:
\begin{equation}
\begin{aligned}
% \begin{cases}
    q_v^t &= \texttt{SA}(F_v^{t}) \in \mathbb{R}^{32\times d}, \\
    % \hat{q_v}^{t} &=\mathcal{P}_{T5} \Big ( \text{Q-Former}(q_v^t, q_l) \Big ) \in \mathbb{R}^{32\times d'},
    \hat{q_v}^{t} &= \text{Q-Former}(q_v^t, q_l) \in \mathbb{R}^{32\times d'},
% \end{cases}
\end{aligned}
\label{eq:q_former}
\end{equation}
where $q_v^t$ and $\hat{q_v}^{t}$ denote the video tokens before and after Q-former at timestamp $t$,
% $\mathcal{P}_{T5}$ is the linear projection in FlanT5, 
$F_v^{t} \in \mathbb{R}^{HW \times d}$ represents the video frame feature at timestamp $t$ generated from the visual encoder~(\textit{i.e.}, EVA \cite{fang2023eva}), $q_l \in \mathbb{R}^{32\times d}$ is a group of learnable embeddings used to transform video content into textual information~\cite{li2023blip2}, and we use $d$ and $d'$ to denote the dimension of visual and textural embeddings, respectively. 
$\texttt{SA}(\cdot)$ is the Slot Attention module~\cite{locatello2020object} to acquire visual representations while further reducing redundancy. 
%
% The ablation study listed in Tab.~\cref{tab:ablation} indicates that slot attention proves to effectively reduce dimensionality with negligible impact on performance. 

\smallskip
\label{sec:encoding}
\noindent \textbf{Timestamp encoding.}
%
% To enhance comprehension of the temporal correlations among video frames and effectively engage with time-specific queries, we incorporate the encoding of timestamps.
%
Conventional techniques like sinusoidal~\cite{vaswani2017attention} or learnable encoding~\cite{dosovitskiy2020image} of timestamps mismatch with language models since they're from different domains.
In contrast,  we propose to transform timestamps into the form of text.
%
% The timestamps associated with each frame of a video are first quantized into integer values and subsequently transformed into natural language descriptions.
%
Subsequently, we leverage the FlanT5 \cite{2020t5} encoder for generating embeddings aligned with temporal information contained in the input text queries. 
%
% In contrast to conventional techniques such as sinusoidal or fully learnable time coding, our approach skillfully circumvents the challenge of aligning temporal encoding with text embedding space. 
%
Our approach skillfully circumvents the challenge of aligning temporal encoding with the text embedding space. 

\smallskip
\noindent \textbf{Adaptive selection via token bank.}
% {\color{red} need some explanations on hard/soft selection}
%
The key to token selection lies in enabling the model to comprehend timestamps and video content, 
% thereby facilitating the identification of tokens within the temporal series that contribute most to a judicious response to the input query.
thereby identifying the most relevant tokens to the given prompt within the time series.
%
% The straightforward approach of selecting the Top-K tokens based on vector similarity, also known as hard selection, can result in information loss when incorrect choices are made.
%
In pursuit of this, we introduce the token bank module, which leverages the weighted aggregation of tokens to dynamically preserve both query-specific and overall contextual information.
Specifically, we initiate the process by creating a set of learnable queries, represented as $q_i \in \mathbb{R}^{n\times d'}$.
% , which are designed to extract pertinent information from the token bank.
%
Employing a cross-attention mechanism, these learnable queries effectively comprehend the input prompt, with the concatenation of timestamps and visual embeddings serving as keys within the cross-attention module.
Meanwhile, the learnable queries play the role of extracting the most relevant visual features $E_{\text{vis}}$ from a long-time series:
% with learnable queries serving as the queries to extract the most context-related visual features $E_{\text{vis}}$ from a long-time series:
\begin{equation}
    \begin{aligned}
    % \begin{cases}
    q_{\text{mid}} &= \texttt{MHCA}\Big [ q_i, \text{T5}_{\text{Enc}}(T_{p}), \text{T5}_{\text{Enc}}(T_{p})\Big ], \\
    E_{\text{vis}} &= \texttt{MHCA}\Big [ q_{\text{mid}}, \texttt{concat}\left(\hat{q_v}, \text{T5}_{\text{Enc}}(T_{t})\right), q_v \Big ],
    % \end{cases}
    \end{aligned}
\label{eq:mem_bank}
\end{equation}
where $T_p$ and $T_t$ represent the text prompt and timestamp, respectively. $q_v$ and $\hat{q_v}$ correspond to the entire video representation before and after Q-former, while $q_{\text{mid}}$ serves as an intermediate token that incorporates textual prompt.
% are the tokens of the entire video in the text domain and image domain, and $q_{\text{mid}}$ is the intermediate token, absorbing the textual prompt information.
$\texttt{MHCA}(\cdot)$ denotes multi-head cross attention and $T5_{\text{Enc}}$ is the FlanT5 encoder \cite{2020t5}.

The selected visual features $E_{\text{vis}}$ will then be processed by Q-former and fed into the language model as the visual embedding.
As queries and keys in the cross attention are aligned within the textual domain, our approach effectively identifies and extracts moment- and content-specific visual representations. 
A more detailed illustration of the pipeline is given in the supplementary materials.
% Appendix \textcolor{red}{B}.

% %
% Then the selected visual features are processed by Q-former, and fed into the language model.
% %
% By interacting with cross-attention under the textual domain, our approach succeeds in the identification and extraction of the most pertinent moment-specific and content-specific representations.
% %
% The detailed pipeline is illustrated in \cref{sec:token_selection}.

\section{Experiments}

The fine-tuning datasets of all ten tasks are built upon nuScenes~\cite{caesar2019nuscenes} and Ego4D~\cite{grauman2022ego4d}.
Additional information (annotations, dataset statistics, implementation details, training strategies, \textit{etc.}) is provided in the supplementary materials.
% Appendix \textcolor{red}{B}.
%\textcolor{red}{in \cref{sec:implementation_details}}.
% (Appendix \textcolor{red}{B}).

% \paragraph{Fine-tune Datasets.} 
% %
% Our fine-tuning datasets corresponding to all tasks are built upon a wide scope of popular datasets, including nuScenes~\cite{caesar2019nuscenes}, DriveLM~\cite{sima2023drivelm}, OpenLane-V2~\cite{wang2023openlanev2} and Ego4D~\cite{grauman2022ego4d}. 
% % Due to space limit, the full suite of protocols, some ablations, and visualizations are provided in the Supplementary.

\paragraph{Evaluation metrics.} 
For localization-related tasks, \textit{i.e.}, Tracking, Box Detection, and Box Prediction, we propose metrics specifically designed for the assessment of VLMs in the context of these tasks.
% One prediction would be considered correct if two prerequisites are met: firstly, the Euclidean distance between the predicted and ground truth box centers on the ground plane must adhere to a predefined threshold; secondly, the predicted object category be accurate.
To be considered a correct prediction, the Euclidean distance between the predicted and ground truth box centers must be within a threshold, and the predicted category should also be accurate.
Mathematically, this can be expressed as: 
\begin{equation}
    \text{Pr}\text{@}\text{k} = \frac{1}{N} \sum_{i=1}^{N} \mathbbm{1} \Big(\Vert \hat{b}^{i} - b_{gt}^{i} \Vert_2 < k \bigcap (\hat{c}^{i} = c_{gt}^{i})\Big),
\label{eq:metrics}
\end{equation}
where $N$ is the number of QA pairs, $\hat{b}^{i}$ and $\hat{c}^{i}$ denote the predictions for box center and object category corresponding to their annotation $b_{gt}^{i}$ and $c_{gt}^{i}$, $\mathbbm{1}$ is the indicator function, and $k$ is the predefined distance threshold.
%
% In our experiments, we set the threshold values as 1 or 2, with Pr@1 serving as the primary metric for assessing performance across the three tasks.
We set Pr@1 as the primary metric in the following experiments.
% for these tasks.
% for assessing performance on localization-related tasks.
%

Regarding the seven language-related tasks, we employ three established metrics, namely CIDEr~\cite{vedantam2015cider}, ROUGE-L~\cite{lin2004rouge}, and BLEU~\cite{papineni2002bleu}.
%, for evaluating the quality and correctness of natural language descriptions.
%
In contrast to the simplistic word-wise evaluation of BLEU and ROUGE-L, CIDEr assesses sentences based on content and semantics, aligning more closely with human judgment \cite{aafaq2019video}. Hence we employ CIDEr as the primary metric for evaluating the quality and correctness of output sentences. 
For the convenience of comparison, a rescaling involving $log_{10}(\text{CIDEr}+1)$ is employed to standardize CIDEr values within the range of 0 to 1.
In addition, we present the aggregate metric for BLEU by averaging values across BLEU-1 to BLEU-4.

% Recognizing the inherent limitations of individual metrics, we utilize this combination to achieve a comprehensive and well-rounded evaluation across various tasks. 
% We use the arithmetic mean of the different metrics as an holistic evaluation metric for a model's performance on a given task.

%##################################################################################################
% overall table
\begin{table*}[t]
%\vspace{-.2em}
\centering
% \small
%################################################# 
% nuScenes
%#################################################
\subfloat[
\textbf{nuScenes.} ELM outperforms the leading previous methods on the matority of metrics across all six tasks on nuScenes, which validates the generality of our model.
%\label{tab:}
]{
%\begin{minipage}{1.0\linewidth}{
\scalebox{0.63}{
\tablestyle{2.0pt}{1.05}
\begin{tabular}{l|x{24}x{24}|x{27}x{27}|x{27}x{27}|x{24}x{24}x{24}|x{28}x{28}x{28}|x{24}x{24}x{24}}
\toprule
\multirow{2}{*}{Methods} & \multicolumn{2}{c|}{Tracking} & \multicolumn{2}{c|}{Box Detection} & \multicolumn{2}{c|}{Box Prediction} & \multicolumn{3}{c|}{Traffic Sign Inquiry} & \multicolumn{3}{c|}{Surrounding Narration} & \multicolumn{3}{c}{Action \& Decision} \\
 & \baseline{Pr@1} & Pr@2 & \baseline{Pr@1} & Pr@2 & \baseline{Pr@1} & Pr@2 & \baseline{C} & R & B & \baseline{C} & R & B & \baseline{C} & R & B\\
\cmidrule{1-16}
 BLIP2-opt \cite{li2023blip2} &  \baseline{0.1} & 0.1 & \baseline{0.1}& 0.2 & \baseline{0.2} & 0.5& \baseline{23.0}  & 26.9 & 20.5& \baseline{8.1} & 19.7 & 21.2 & \baseline{8.4} & 11.5  & 11.1 \\
 BLIP2-flant5 \cite{li2023blip2} & \baseline{3.0} & 6.0 & \baseline{5.1}& 10.5 & \baseline{3.6} & 6.3& \baseline{63.1} & 39.4& 31.4 & \baseline{65.2} & 64.9 & 27.9 & \baseline{68.7} & 71.4 & 43.1 \\
 % Flamingo \cite{alayrac2022flamingo}& \baseline{-} & - & \baseline{-}& - &  \baseline{-} & -& \baseline{-} & - & -& \baseline{-} & - & -& \baseline{-} & - & - \\
 LLaMA-Ada. \cite{gao2023llamaadapterv2} & \baseline{6.1} & 10.5 & \baseline{8.3}& 14.9 & \baseline{7.5} & 12.5& \baseline{\underline{68.3}} & \underline{66.6} & \underline{61.6} & \underline{\baseline{67.0}} &  \underline{77.5} & \textbf{60.1}& \baseline{\underline{72.3}} & \underline{76.8} & \textbf{64.7} \\
 LLaVA \cite{liu2023llava} & \baseline{5.5} & 9.3 & \baseline{28.5} & 31.2 & \baseline{6.1} & 10.2& \baseline{51.1} & 58.5 & 50.8 & \baseline{64.9} & 64.6 & \underline{41.2} & \baseline{64.4} & 62.4 & \underline{57.9}\\
 Otter \cite{li2023mimicit} & \baseline{\underline{10.0}} & \underline{17.2} & \baseline{\underline{41.8}} & \underline{46.9} & \baseline{\underline{8.9}} & \underline{15.8}& \baseline{62.8} & 41.1 & 32.4 & \baseline{60.0} & 64.2 & 13.3 & \baseline{69.2} & 73.2 & 53.0\\
 VideoChat \cite{2023videochat}& \baseline{0.4} & 0.9 & \baseline{0.1}& 0.3 & \baseline{0.1} & 0.2& \baseline{25.3} & 21.9 & 11.7& \baseline{21.7} & 29.2 & 12.2& \baseline{29.6} & 33.2 & 13.1 \\
 Vid-ChatGPT \cite{Maaz2023VideoChatGPT} & \baseline{0.1} & 0.6 & \baseline{0.1}& 1.0 & \baseline{0.3} & 1.2& \baseline{49.6} & 57.1 & 48.6 & \baseline{61.0} & 69.6 & 37.2 & \baseline{53.6} & 58.5 & 43.5 \\
 \midrule
 \textbf{ELM (Ours)}& \baseline{\textbf{14.0}} & \textbf{23.3} & \baseline{\textbf{51.6}}& \textbf{56.9} & \baseline{\textbf{15.1}}  & \textbf{24.4}& \baseline{\textbf{76.5}} & \textbf{71.2} & \textbf{63.9} & \baseline{\textbf{73.2}} & \textbf{78.7} & 29.8 & \baseline{\textbf{74.4}} & \textbf{83.3} & 41.2 \\
 \bottomrule
\end{tabular}
}
}
%\end{minipage}}
\vspace{1pt}
%#################################################
% Ego4D 
%#################################################
% \hspace{2pt}
\subfloat[
\textbf{Ego4D.} We extend the model to Ego4D dataset and verified the generality of our token bank module on four tasks.
%\label{tab:}
]{
%\begin{minipage}{0.72\linewidth}{
\tablestyle{5.0pt}{1.05}
\scalebox{0.63}{
\begin{tabular}{l|x{16}x{16}x{16}|x{14}x{14}x{14}|x{16}x{16}x{16}|x{20}x{20}x{20}}
\toprule
\multirow{2}{*}{Methods} & \multicolumn{3}{c|}{Moment Recap} & \multicolumn{3}{c|}{Event Query} & \multicolumn{3}{c|}{Ego. Narration} & \multicolumn{3}{c}{Activity Prediction} \\
 & \baseline{C} & R & B & \baseline{C} & R & B & \baseline{C} & R & B & \baseline{C} & R & B \\
\cmidrule{1-13}
 BLIP2-opt \cite{li2023blip2} & \baseline{1.2} & 8.9 & 6.8  & \baseline{7.8} & 28.4 & 14.7& \baseline{5.2} & 19.8 & 10.7 & \baseline{2.7} & 18.7 & 9.6 \\
 BLIP2-flant5 \cite{li2023blip2} & \baseline{13.1} &  31.9 & 12.5 & \baseline{27.3} & 33.0 & 16.6 & \baseline{16.9} & 33.5 & 15.4 & \baseline{11.5} & 31.2 & 11.3 \\
 % Flamingo \cite{alayrac2022flamingo} & \baseline{-} & - & -& \baseline{-} & - & -& \baseline{-} & - & -& \baseline{-} & - & - \\
LLaMA-Ada. \cite{gao2023llamaadapterv2} & \baseline{11.2} & 30.2 & 12.3& \baseline{37.5} & \textbf{47.2} & 28.1& \baseline{18.4} & 34.2 & 15.3& \baseline{\underline{13.1}} & 31.2 & 12.8 \\
 LLaVA \cite{liu2023llava} & \baseline{9.6} & 28.3 & 12.1 & \baseline{\textbf{39.8}} & \underline{44.6} & \textbf{29.9} & \baseline{6.5} &  28.2 & 11.6 & \baseline{8.4} & 28.0 & 13.0 \\
 Otter \cite{li2023mimicit} & \baseline{11.4} & 29.6 & 10.5 & \baseline{27.1} & 38.3 & 19.1 & \baseline{14.1} & 31.4 & 13.9 & \baseline{11.1} & 29.4 & 10.3 \\
 VideoChat \cite{2023videochat}& \underline{\baseline{13.2}} & \underline{32.5} & \underline{13.8} & \baseline{34.5} & 42.2 & 26.4& \baseline{\underline{20.7}} & \underline{35.0} & \textbf{17.6} & \baseline{12.1} & \underline{32.4} & \underline{14.1} \\
 Vid-ChatGPT \cite{Maaz2023VideoChatGPT}& \baseline{10.0} & 31.1 & 13.3 & \baseline{27.9} & 36.5 & 20.9 & \baseline{10.2} & 21.7 & 10.4 & \baseline{9.4} & 30.5 & 12.6\\
 \midrule
 \textbf{ELM (Ours)}& \baseline{\textbf{22.6}} & \textbf{36.7} & \textbf{19.4}& \baseline{\underline{38.0}} & 43.1 & \underline{27.6} & \baseline{\textbf{26.5}} & \textbf{37.7} & \underline{16.9} & \baseline{\textbf{18.1}} & \textbf{34.1} & \textbf{17.0}\\
 \bottomrule
\end{tabular}
%}\end{minipage}
}
}
% \hspace{-0.5mm}
%#################################################
% params
%#################################################
\subfloat[
\textbf{Parameters.}
%\label{tab:}
]{
%\begin{minipage}{0.12\linewidth}{
\tablestyle{4.0pt}{1.05}
\scalebox{0.63}{
\begin{tabular}{l|c}
\toprule
\multirow{2}{*}{Methods} & \multirow{2}{*}{Param.}\\
& \\
\cmidrule{1-2}
 BLIP2-opt  & 2.7B \\
 BLIP2-flant5  & 2.7B \\
 % Flamingo & 3B \\
LLaMA-Ada.  & 7B \\
 LLaVA  & 7B \\
 Otter & 7B \\
 VideoChat & 7B \\
 Vid-ChatGPT  & 7B \\
 \midrule
 \textbf{ELM (Ours)} & 2.7B \\
 \bottomrule
\end{tabular}
%}\end{minipage}
}
}
% \vspace{-.3em}
\caption{\textbf{Comparison to State-of-the-arts}. All methods are \textbf{fine-tuned} on the corresponding tasks.
The main metrics (\%) are marked in \colorbox{baselinecolor}{gray}.
\textbf{Bold} emphasizes top method; \underline{underline} marks the runner-up.
\texttt{C}: CIDEr; \texttt{R}: ROUGE-L; \texttt{B}: BLEU.
}
% \vspace{-10pt}
\label{tab:sota} 
\underfigtab
\end{table*}

\begin{figure*}[t]
    \centering
    \includegraphics[width=\linewidth]{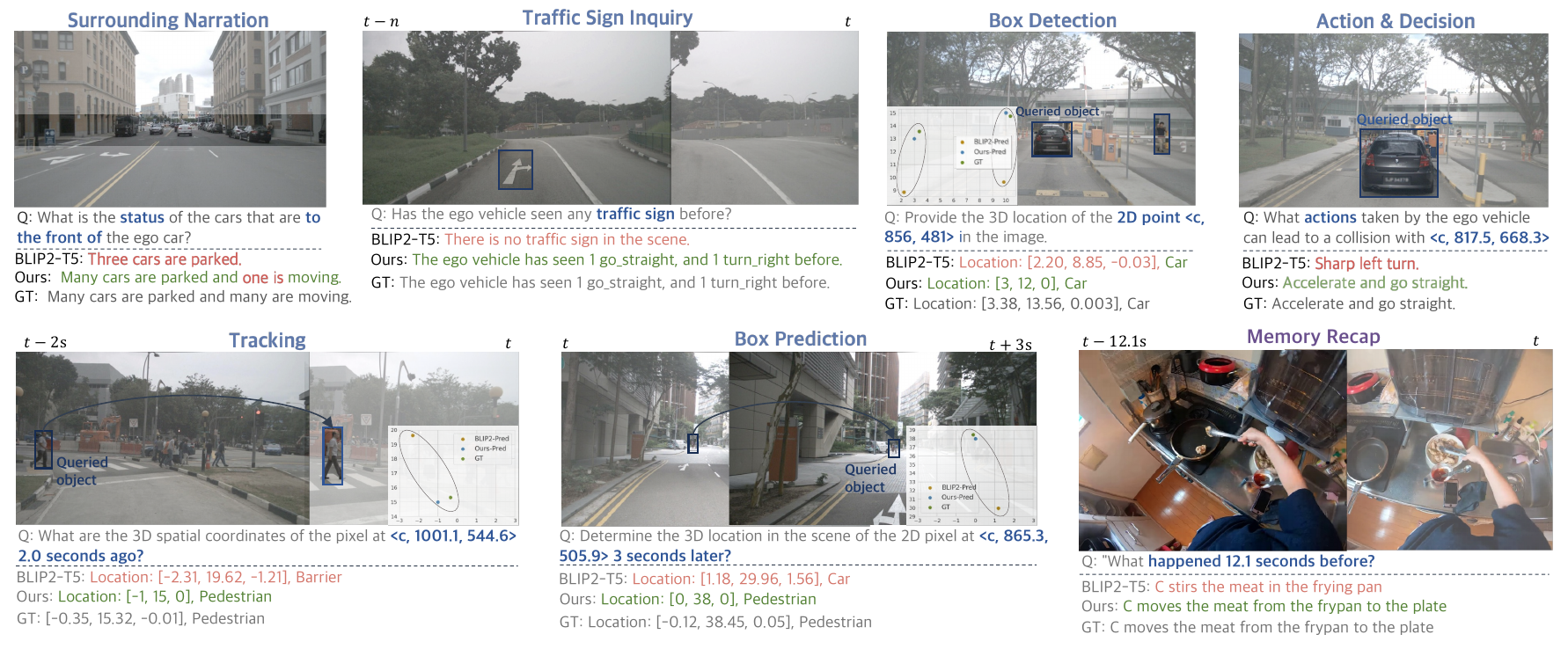}
    \vspace{-15pt}
    \caption{\textbf{Visualization on 
    % embodied understanding 
    the
    benchmark.} 
    We provide results for seven tasks through images and corresponding QA pairs. 
    % Due to page limit, the 
    % three 
    The remaining tasks are included in the Appendix.
    % In comparison to the baseline of BLIP2-flant5 \cite{li2023blip2}, our model is superior in terms of consistency with ground truth.
    }
    \vspace{5pt}
    \label{fig:vis}
    \underfigtab
    % \vspace{5pt}
\end{figure*}

\subsection{Comparison to State-of-the-arts}

We first show the performance comparison of ELM and previous state-of-the-art VLMs \cite{li2023blip2,alayrac2022flamingo,gao2023llamaadapterv2,liu2023llava,li2023mimicit,2023videochat,Maaz2023VideoChatGPT} on our proposed benchmark.
All VLMs are initialized using the official pre-trained weights and then fine-tuned on our dataset.
% All VLMs are fine-tuned using the fine-tuning datasets and initialized with the officially provided pre-trained weights.
%\cref{fig:radar} illustrates the noteworthy advantages of ELM over the previous top-performing model in specific tasks, establishing our method as a frontrunner overall.
%
Detailed metrics with respect to all tasks are documented in \cref{tab:sota}. 
On localization-related tasks such as Box Detection,
our model attains a significant superiority.
Notably, our method surpasses Otter~\cite{li2023mimicit} with a remarkable margin of \textbf{+9.8\%} in Pr@1 score, illustrating the effectiveness of our proposed space-aware pre-training.
On time-related tasks, \textit{e.g.}, Traffic Sign Inquiry and Moment Recap, we surpass the second-best by \textbf{+13.4\%} and \textbf{+6.8\%} in CIDEr score, respectively.
This highlights ELM's outstanding ability in retrieving timestamp information, attributed to the time-aware token selection.
% underscoring the significance of time-aware token selection for augmenting temporal competency.
%
We notice that LLaVA \cite{liu2023llava} exhibits superior performance compared to ELM in Event Query task that focuses on successive event reasoning. 
%
% This performance difference can be attributed to the limitations of ELM due to the inherent logical understanding capabilities of the FlanT5 model~\cite{2020t5}. 
% These limitations may restrict ELM's ability to effectively handle this specific task.
ELM, which excels in precise timestamp retrieval, may face limitations in handling this specific task due to the inherent constraints in the FlanT5~\cite{2020t5} model's capacity to comprehend lengthy texts.
% In contrast, ELM excels the precise timestamp information retrieval and is constrained by the logical comprehension capacity inherent in the FlanT5 model~\cite{2020t5}, potentially leading to limitations in addressing this specific task.
%
Besides, due to the preference of our model for generating concise responses, its performance in terms of BLEU is affected \cite{aafaq2019video}.
\cref{fig:vis} demonstrates the qualitative comparison between ELM and baseline method~(\textit{i.e.}, BLIP2-flant5 \cite{li2023blip2}) on nuScenes \cite{caesar2019nuscenes} and Ego4D \cite{grauman2022ego4d} dataset. 
It is observed that ELM's output is much closer to the ground truth, especially in tasks involving 3D localization.
Additional visualizations are shown in the supplementary materials.
% Appendix \textcolor{red}{C}.

\begin{table*}[t]
% \vspace{-2em}
\centering
%#################################################
% Ablations on pre-training
%#################################################
\subfloat[
\textbf{Ablations on pre-training.}
]{

\begin{minipage}{0.55\linewidth}{
\begin{center}
\tablestyle{2.0pt}{1.0}
% \hspace{-1cm}
\scalebox{0.73}{
\begin{tabular}{l|cc|ccc|cccc}
\toprule
 & Vocab & Data &  T & BD & BP & TSI & SN & AD & EN  \\ \midrule
0 & - & -  &    3.0     &    5.1     &   3.6      &    63.1  &        65.2     &       68.7     &    16.9       \\ \midrule
1 & Rewritten & - & 2.8 & 5.9  &  3.0  & - & - &   -     &  -             \\ 
2 & - & Loc  &  6.5	 & 31.2	 &  7.7  & - &  - & - &    -       \\
3 & Rewritten & Loc & 12.2  & 46.5  &   12.6  &  59.4 & 63.7 &  63.8   & 16.2           \\ \midrule
\baseline{\textbf{7}}& \baseline{Rewritten} & \baseline{Des, Loc} & \baseline{\textbf{14.0}} & \baseline{\textbf{51.6}}       & \baseline{\textbf{15.1}}         & \baseline{\textbf{76.5}}     & \baseline{\textbf{73.2}}     & \baseline{\textbf{71.4}}      & \baseline{\textbf{26.5}}       \\ \bottomrule
\end{tabular}
}
\end{center}
}
\end{minipage}
}
%\hspace{1mm}
%#################################################
% Ablations on token selection.
%#################################################
\subfloat[
\textbf{Ablations on token selection.} 
]{
\begin{minipage}{0.4\linewidth}{\begin{center}
\tablestyle{2.0pt}{1.0}
% \hspace{-0.5cm}
\scalebox{0.73}{
\begin{tabular}{l|cc|ccc}
\toprule
 & Encoding & Selection &  MR & EQ & AP \\ \midrule
0 & - & - & 13.1    &  27.3 &   11.5\\ \midrule
4  & Sinusoidal &  -  &   12.3   &  34.8   &    12.1    \\ 
% (h)&   hard sel.  &  - & -  &   -  & -&- &  -      &     -       &      &     &        \\ 
5& Textual &   Hard  &  17.8    &   37.3  &    13.3    \\ 
6 & - &   Manual   &    18.9   &  \textbf{39.4}    &    17.6     \\ \midrule
\baseline{\textbf{7}}& \baseline{Textual} & \baseline{Soft}  & \baseline{\textbf{22.6}}           & \baseline{38.0}          & \baseline{\textbf{18.1}}       \\ \bottomrule
\end{tabular}
}
% \hspace{-10mm}
\end{center}
}
\end{minipage}
}
\hspace{0.5mm}
%#################################################
% Comparisons on 3D detection.
%#################################################

\vspace{-5pt}
\caption{
\textbf{
% Detailed 
Ablations on the effectiveness of each component.} 
Baseline (Exp.0) uses the BLIP2-flant5 \cite{li2023blip2} model. 
ELM (Exp.7) is marked in \colorbox{baselinecolor}{gray}. We only show the main metrics for brevity.
% We give an explanation for the acronym here.
% Hongyang: remove the "." in the abbreviation.
\texttt{T}: Tracking, \texttt{BD}: Box Detection, \texttt{BP}: Box Prediction, \texttt{TSI}: Traffic Sign Inquiry; \texttt{SN}: Surrounding Narration; \texttt{AD}: Action \& Decision; \texttt{EN}: Egocentric Narration; \texttt{MR}: Moment Recap; \texttt{EQ}: Event Query; \texttt{AP}: Activity Prediction; \texttt{Loc}: Localization; \texttt{Des}: Description; \texttt{C}: CIDEr; \texttt{R}: ROUGE-L; \texttt{B}: BLEU.
% ``BF.'': BEVFormer \cite{li2022bevformer}; ``BD.'': BEVDepth \cite{li2023bevdepth}.
}
% \vspace{-20pt}
\label{tab:ablation}
\underfigtab
\end{table*}
% \hspace{-18pt}

% \hspace{-15pt}
\begin{figure}
\begin{minipage}[]{0.55\linewidth}{\begin{center}
\vspace{0pt}
\tablestyle{2.0pt}{1.0}
% \hspace{-0.2cm}
\scalebox{0.73}{
\setlength{\tabcolsep}{1.5mm}{
\begin{tabular}{l|cc|c|c}
\toprule
 Method & $A_{\text{GPT}}$ & $S_{\text{GPT4V}}$ & $D_{\text{n-gram}}$ & Time(s/\#)$\downarrow$ \\ \midrule
 Baseline & 54.3 & 34.4 & 14.8 & \textbf{1.6}\\
 + Filtering & 68.3 & 49.5 & 21.2 & 1.9\\
 \baseline{+ Verification} & \baseline{\textbf{84.4}} & \baseline{\textbf{66.9}} & \baseline{\textbf{26.7}} &  \baseline{4.5}\\ \midrule
 \textit{Manual Labeling} & \textit{100} & \textit{64.3} & \textit{23.3} & \textit{72.4} \\
\bottomrule
\end{tabular}
}
}
\end{center}
}
\makeatletter\def\@captype{table}\makeatother\caption{\textbf{Labeling quality and corresponding time cost.} \texttt{Baseline}: LLaMA-Ada., \texttt{$A_{\text{GPT}}$}: 
accuracy between auto and manually annotated text evaluated by GPT, \texttt{$S_{\text{GPT4V}}$}: rationality score in image-text matching evaluated by GPT4V, \texttt{$D_{\text{n-gram}}$}: 
diversity evaluated by distinct n-gram ratio of phrases. 
Time refers to the average duration required for a single person to annotate a piece of data.
Our choice is marked in \colorbox{baselinecolor}{gray}.
}
\label{tab:data_quality}
\end{minipage}
\hspace{1pt}
\begin{minipage}[]{0.43\linewidth}{\begin{center}
\vspace{-2pt}
\tablestyle{2.0pt}{1.0}
% \hspace{-0.2cm}
\scalebox{0.73}{
\setlength{\tabcolsep}{1.5mm}{
\begin{tabular}{l|cc|c}
\toprule
Method & ADE$\downarrow$  & FDE$\downarrow$ & Time(s)$\downarrow$ \\ \midrule
Command Mean & 7.98 & 11.41 & - \\
UniAD-single \cite{hu2023_uniad} & 4.16 & 9.31 & 0.56 \\
Flamingo \cite{alayrac2022flamingo} & 2.78 & 5.31 &  1.47\\ \midrule
\baseline{ELM} & \baseline{\textbf{2.28}} & \baseline{\textbf{4.27}} & \baseline{1.61} \\ \bottomrule
\end{tabular}
}
}
\end{center}}
    \makeatletter\def\@captype{table}\makeatother\caption{\textbf{Planning on out-of-distribution datasets.} Command Mean denotes the average value of the trajectories corresponding to each instruction in the training set.
\texttt{ADE} \& \texttt{FDE}: average \& final distance error (m) of future trajectory in 3 seconds. 
All methods are trained on nuScenes \cite{caesar2019nuscenes} and evaluated on Waymo \cite{sun2020scalability}. 
}
\label{tab:ood_planning}
\end{minipage}
\vspace{-10pt}
\end{figure}

%
% {\color{red}It is worth noting that the introduced description data improves the model's performance on Tracking, Box Detection, and Box Prediction by \textbf{+1.8\%}, \textbf{+5.1\%}, \textbf{+2.5\%}, suggesting that descriptive QA has the potential of boosting model's understanding on spatial location relationships.}
%
\subsection{Ablation Study}

% \noindent \textbf{Effectiveness of each proposed component.}
%
We conduct ablation studies to assess the effectiveness of each component, with experiments shown in~\cref{tab:ablation}. Exp.0 serves as a baseline built upon BLIP2-flant5 \cite{li2023blip2} and Exp.7 represents the final design of ELM.
% The best result of each metric is marked in bold, and the runner-up result is underlined in each column.
%
Initially, we examine the pre-training strategy within our pipeline in \cref{tab:ablation} (a). 
Comparative analysis between Exp.0 and Exp.2 reveals that solely performing localization pre-training without rewriting the space-relevant vocabulary yields limited improvements.
Notably, the collaborative application of vocabulary rewriting and localization pre-training manifests a substantial advancement across all three localization tasks, exemplified by improvements of \textbf{+9.2\%}, \textbf{+41.4\%}, and \textbf{+9.6\%} in Tracking, Box Detection, and Box Prediction, respectively.
% In Exp. 1-3, we show the cooperative effect of rewriting vocabulary and pre-training on localization data, with an improvement of +9.2/+41.4/+9.5 on AP@1.
% The performance on localization gets improved when rewritten vocabulary and large-scale training are introduced simultaneously (Exp.3, +9.2/+41.4/+9.5 on AP@1).
% 
Nevertheless, a decrement in performance on alternative tasks is observed in Exp.3, prompting the adoption of cooperative pre-training in our final configuration~(Exp.7), which brings enhanced performance across all tasks.
This underscores the significance of integrating both localization and description data during the pre-training phase.
We note that the model's performance in the localization-related tasks improves by \textbf{+1.8\%}, \textbf{+5.1\%}, and \textbf{+2.5\%} after the incorporation of descriptive data.
We believe this is due to the fact that descriptive data provides information about relative positional relationships that benefits the localization tasks.

% {\color{red}[todo] Lack explanation about 'Hard' / 'Soft?} 
In addition, we explore several implementations of the token selection module, as results listed in \cref{tab:ablation} (b).
The utilization of a straightforward sinusoidal temporal encoding may result in a marginal performance decline, potentially stemming from the model's difficulty in interpreting temporal information in this encoding scheme.
It is worth noting that hard selection denotes selecting the tokens of three frames with the highest attention scores, while soft selection is the weighted summation across all tokens.
Manual selection, which involves picking the tokens of the three frames based on the ground truth timestamp, is the theoretically optimal solution to hard selection.
Using textual encoding strategy~(as detailed in \cref{sec:encoding}) with hard selection results in a noticeable improvement of \textbf{+4.7\%}, \textbf{+10.0\%}, and \textbf{+1.8\%} on three tasks.
% will discard all information of unselected tokens, so there will be serious performance degradation when picking the wrong ones.
%
Ultimately, we incorporate soft token selection, potentially encompassing information across all tokens, into our model. 
This adaptation brings improved performance on Moment Recap and Activity Prediction tasks, denoted as \textbf{+9.5\%} and \textbf{+6.6\%}, respectively, while preserving comparability with manual selection on the Event Query task.

\subsection{Further Discussions and Analysis}
\label{sec:discussion}

% Appendix \textcolor{red}{C}.
% {\color{red} todo: explanation of the two metrics, including why they are unfair}
\begin{figure}[t]
% \hspace{-18pt}
\begin{minipage}[t]{0.57\linewidth}
    \centering
{\begin{center}
\tablestyle{2.0pt}{1.0}
% \hspace{-0.2cm}
\scalebox{0.73}{
\begin{tabular}{l|ccccc}  
\toprule
Method & \baseline{$\text{Pr}^{*}@1$}  & $\text{Pr}^{*}_{{car}}$@1 & $\text{Pr}^{*}_{{ped}}$@1 & $\text{Pr}^{*}_{{bar}}$@1 & $\text{Pr}^{*}_{{tra}}$@1 \\  \midrule 
DETR3D \cite{wang2022detr3d}  & \baseline{43.6} & 48.9 & 44.6  & 39.1 & 15.6\\ 
BEVFormer \cite{li2022bevformer}  & \baseline{47.4} & 52.3 & 48.8 & 43.5 & 14.3 \\ 
VCD \cite{huang2023leveraging}   & \baseline{53.4} & 50.3 & 60.0 & 68.1 & 20.6\\
\midrule \midrule
% \hline
Method & \baseline{$\text{Pr}@1$}  & $\text{Pr}_{{car}}$@1 & $\text{Pr}_{{ped}}$@1 & $\text{Pr}_{{bar}}$@1 & $\text{Pr}_{{tra}}$@1 \\
\midrule 
\textbf{Ours}  & \baseline{51.6} & 64.9 & 50.2 & 70.4 & 26.4 \\
\bottomrule
\end{tabular}
}
\end{center}
}
    \makeatletter\def\@captype{table}\makeatother\caption{
\textbf{Extended evaluation on 3D detection performance.}
The Hungarian algorithm \cite{kuhn1955hungarian} is employed to ensure a reasonably fair comparison between ELM and conventional 3D detection models.
\texttt{ped}: pedestrian; \texttt{bar}: barrier; \texttt{tra}: trailer.
The main metric is marked in \colorbox{baselinecolor}{gray}.
% \texttt{MR}: Moment Recap; \texttt{EQ}: Event Query; \texttt{AP}: Activity Prediction.
 }
\label{tab:3d_det}
\end{minipage}
\hspace{2pt}
\begin{minipage}[t]{0.4\linewidth}
    \centering
    {\begin{center}
    \tablestyle{2.0pt}{1.0}
    % \hspace{-0.2cm}
    \scalebox{0.73}{
    \begin{tabular}{l|c|ccc}  
    \toprule
     Task& Method & Pr@1 & Pr@2 & Pr@4 \\  \midrule 
    \multirow{2}{*}{Tracking}   & SFT & \textbf{14.0} & \textbf{23.3} & \textbf{36.9} \\ 
       & Zero-shot & 9.8 & 14.8 & 23.0 \\ \midrule \midrule
    Task & Method & C & R & B \\  \midrule 
    Action  & SFT & \textbf{71.4} & \textbf{74.6} & \textbf{43.0} \\
    \& Decision & Zero-shot & 59.0 & 65.0 & 35.3  \\
    \bottomrule
    \end{tabular}
    }
    \end{center}}
    \makeatletter\def\@captype{table}\makeatother\caption{\textbf{Zero-shot evaluations on new tasks.} Our model is also capable of achieving decent performance on zero-shot scenarios in comparison to supervised fine-tuning~(SFT).
    }
\label{tab:ood}
\end{minipage}
\vspace{-10pt}
\end{figure}

\smallskip \noindent \textbf{Evaluation on label quality and diversity.}
To verify the reliability of the auto-labeling pipeline, we manually annotate thousands of images and conduct a quantitative experiment for auto-labeling in \cref{tab:data_quality}.
The baseline is an auto-labeling pipeline using LLaMA-Adater V2 \cite{gao2023llamaadapterv2}.
$A_{\text{GPT}}$ is the accuracy between automatically annotated text and human-annotated text evaluated by GPT4 \cite{openai2023gpt}, $S_{\text{GPT4V}}$ is the rationality score in image-text matching evaluated by GPT4V, and $D_{\text{n-gram}}$ is diversity, evaluated by different $n$-gram ratios in a phrase.
Results show that our auto-labeling quality nearly equals manual annotation and leads in diversity (\textbf{26.7}), signifying the excellence of our data.
Additionally, we report the average time required to collect each piece of data using different annotation methods. 
Manual image labeling entails meticulous inspection and detailed textual description, demanding significant time investment. Conversely, automated annotation pipelines enable annotators to efficiently filter and rectify errors in sampled image batches, substantially decreasing time consumption.

\smallskip \noindent \textbf{Out-of-distribution evaluation.}
To adequately demonstrate the model's generalization ability,
\cref{tab:ood_planning} shows experiments on planning in unseen data, which includes both temporal and spatial understanding of ego vehicle future trajectory.
Each frame in nuScenes is associated with one of 3 commands: \texttt{turn\_left}, \texttt{turn\_right}, or \texttt{go\_straight}. 
The baseline, Command Mean, uses the mean of all trajectories in the training set whose command matches the current test frame command.
Moreover, we compare our method with the current state-of-the-art method on nuScenes, UniAD \cite{hu2023_uniad}. In addition to the released checkpoint that requires multi-view input, we trained a single-frame version (UniAD-Single) for a fair comparison with our single-frame VLM.
All methods are trained solely on the front view of nuScenes and are applied to Waymo without fine-tuning or adaptation.
Please refer to the supplementary materials for the detailed design of using ELM for planning.
ELM achieves respectable results in novel scenarios, surpassing end-to-end driving (UniAD) and other VLMs (Flamingo).

\smallskip
\noindent \textbf{Comparison to traditional 3D perception task.}
Addressing concerns pertaining to the superiority of embodied understanding over traditional 3D localization methods, our model is benchmarked against DETR3D \cite{wang2022detr3d}, BEVFormer \cite{li2022bevformer}, and VCD \cite{huang2023leveraging}, as listed in \cref{tab:3d_det}.
Although our QA-based approach does not produce confidence scores, we have made efforts to conduct fair comparisons.
The {Pr}$^{*}$@1 metric is derived from \cref{eq:metrics} after performing a Hungarian algorithm \cite{kuhn1955hungarian} to establish a one-to-one matching between the prediction and the ground truth.
%
% A one-to-one matching is established between the output of the conventional detection model and the ground truth, thus we can calculate the evaluation metric 
% $\text{Pr}^{*}@1$ through \cref{eq:metrics}.
% However, it is important to note that the QA approach lacks confidence scores, which raises concerns about potential unfairness in the evaluation process.
%
The results show that ELM is comparable to classical models in 3D perception. Additional comparisons are in the supplementary materials.

\smallskip
\noindent \textbf{Zero-shot on new tasks.}
To assess the generalization of ELM across different tasks within the benchmark, we fine-tune it using data associated with Box Detection and Moment Recap tasks, with subsequent testing on Tracking. 
Additionally, we fine-tune the model on Surrounding Narration and Activity Prediction, followed by inference on the Action \& Decision task. 
% \cref{tab:ablation} (c) represents the comparison of the model's zero-shot versus supervised learning on new tasks. 
% %
% For zero-shot on the Tracking task, our model is solely trained on the Box Detection and Moment Recap tasks beforehand.
% %
% And for that on Action \& Decision, the model is trained on Surrounding Narration and Activity Prediction. 
%
The results in \cref{tab:ood} indicate that the model's zero-shot capability, to
handle tasks unseen before, is on par with supervised learning. 
%
% This suggests that our model possesses noteworthy generalization and adaptability to novel tasks.
% The results in \cref{tab:ablation} (c) suggest that the model's zero-shot capability in handling previously unencountered tasks is on par with supervised learning, demonstrating the noteworthy generalization capacity of our model and its aptitude for adapting to novel tasks.
Notably, even evaluated in a zero-shot manner, ELM performs comparably to the previous VLM on both tasks (see \cref{tab:sota}). We attain a zero-shot performance of \textbf{9.8\%} in tracking compared to Otter's \textbf{10.0\%}.

\begin{figure*}[t]
    \centering
    \includegraphics[width=\linewidth]{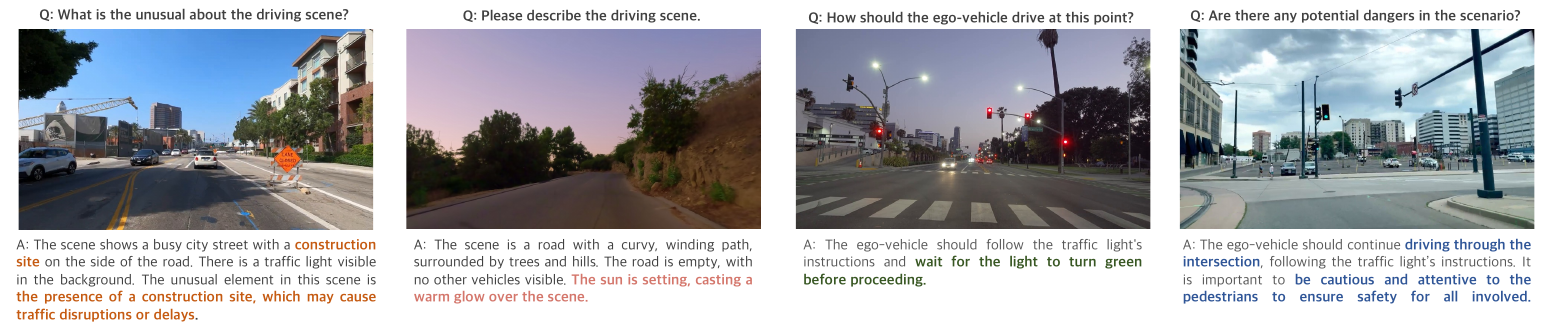}
    \vspace{-22pt}
    \caption{\textbf{Zero-shot on new scenarios.} 
    We select images from the internet that are not utilized during the training to assess the model's proficiency in unexplored scenarios.
    The results validate our model's ability to create notably logical interpretations.
    } 
    \label{fig:zero_shot}
    \underfigtab
\end{figure*}

\smallskip
\noindent \textbf{Verification of open scene understanding.}
We evaluate ELM on novel scenarios and tasks to validate its generalization.
The visualization in 
\cref{fig:zero_shot} demonstrates the superior scene understanding ability of our model on unseen data.
Impressively, it can pay attention to construction signs on the road, make rational driving decisions, and analyze potential dangers. 
This showcases the potential to surpass traditional perception models in understanding unseen scenarios.
% shows the zero-shot scene understanding ability of our model on world data.

\section{Conclusion and Limitation}

We apply VLMs to achieve embodied understanding of driving scenarios and present a benchmark consisting of a suite of tasks and rubrics.
ELM
% , an embodied language model, 
is proposed for the pursuit of understanding driving scenes in long-scope space and time, exhibiting promising generalization performance.

\smallskip
\noindent
\textbf{Limitations and future work.}
% Even though ELM exhibits promising generalization properties, there are multiple limitations of this work.
%
Currently, ELM only perceives driving scenes and interacts with human users. 
%
% In our future work, 
ELM can be further explored to generate driving control signals. 
Additionally, we will implement a prototype system, making ELM an embodied agent for closed-loop autonomous driving.
Further experiments are needed to examine the model's capacity in broader scenarios, as our databases are mostly nuScenes \cite{caesar2019nuscenes} and Ego4D \cite{grauman2022ego4d}.
To promote the adoption of this model in real-world deployments, more validations need to be conducted to verify whether common sense reasoning helps decision-making in novel scenarios.

% \smallskip
% \noindent
% \textbf{Broader impact.}

% {
%     \small
%     \bibliographystyle{ieeenat_fullname}
%     \bibliography{short,main}
% }
\bibliographystyle{splncs04}
\bibliography{main}

\newpage

\noindent\textbf{\large{Appendix}}

\appendix

\section{Motivating Questions}

To better understand our work, we supplement with intuitive questions.
% that one might raise.

\bigskip

\textbf{Q1:} \textit{Why is the embodied understanding necessary for driving scenarios?}
\smallskip

Embodied understanding refers to the ability of an autonomous agent to observe, comprehend, and interact with its environments incorporating sensory input and world knowledge \cite{grauman2022ego4d}.
%, and the understanding of spatial-temporal properties of objects.
%
% In the context of driving scenarios, embodied understanding brings several key benefits.
%
Embodied understanding tasks serve to facilitate the agent's proficiency in common sense reasoning. 
For instance, one such scenario involves deciphering the body language of a pedestrian gesturing for the ego-vehicle to proceed first.
Thus, it can implement more complex and safer driving strategies, capable of adapting to dynamic and unexpected scenarios.
Besides, agents can interact with humans more naturally and offer insights about the users' behaviors and expectations, leading to a more intuitive, responsive, and user-friendly design \cite{mu2023embodiedgpt}.
Moreover, it enables agents to learn from each human interaction and adapt over time.

\bigskip

\textbf{Q2:} \textit{Why should Vision-Language Models (VLMs) be introduced into the embodied understanding of driving scenes?}

\smallskip

One of the critical strengths of VLMs \cite{li2023blip2,driess2023palme,touvron2023llama,openai2023gpt} is that it possesses world knowledge from global data, which can help autonomous vehicles in common sense reasoning \cite{brown2020language, chen2023language}. 
VLMs enable a more thorough interpretation of scenes, such as billboards, landmarks, body language, uncommon objects, and more (please refer to \cref{fig:description_label}). 
For these unusual cases, it is hard for end-to-end models \cite{hu2023_uniad,hu2022stp3} to cover them all, even with the introduction of massive additional data.
Nevertheless, VLMs can effortlessly acquire this knowledge.
Moreover, VLMs have the inherent capacity to correlate and generate contextually relevant sentences or instructions based on visual inputs.

\bigskip

\textbf{Q3:} \textit{Will the timeliness of the VLMs ensure it is adequate for driving scenarios?}

\smallskip

Without optimizations, ELM and other VLMs run about an order of magnitude slower than UniAD \cite{hu2023_uniad} (\textbf{0.62} FPS).
However, optimizations for simulated closed-loop development make practical VLM use in driving possible \cite{sima2023drivelm}.
Techniques like distillation and quantization in LLM inference can help, and another approach would be to execute only the final motion stage of an agent at 20 FPS, while the other VQA stages are executed at a lower frame rate.
In addition, works like MobileVLM \cite{chu2023mobilevlm} and TinyLlama \cite{zhang2024tinyllama} are addressing the issue of deploying VLMs on mobile devices, \textit{e.g.}, vehicles.

%It is equally important to remember that inherent limitations do not necessarily imply utilizing the technology. 
%
%We must fully comprehend these limits and develop approaches to overcome or bypass them. 
%The potential benefits of incorporating VLM into the embodied understanding of driving scenes are worth exploring, and it could enhance the overall performance and safety of autonomous driving systems.

\bigskip

\textbf{Q4:} \textit{Why adapt VLMs to driving rather than adding language inputs to driving-specific models?
}

\smallskip

We apply VLMs for driving instead of utilizing pre-processed information from drive-specific models due to training data issues.
A general VLM can benefit from massive pre-training data extracted from the internet and fine-tuning on small driving datasets for adaptation \cite{lecun2022path}. 
It learns from diverse data sources and potentially generalizes new tasks (please refer to Tab. {\color{red}8} in the main paper).
% It can learn from data sources with different labels and has the potential to generalize to new tasks, which is clarified in Tab. {\color{red}4}(c).
Conversely, drive-specific models can only be pre-trained on limited datasets, and incorporating non-driving language input into these datasets is non-trivial.
Combining the advantages of VLMs and driving-specific models is worth further study.

% \bigskip

% \textbf{Q5:} \textit{Why is it necessary to retrain VLMs on driving scenarios instead of loading and fine-tuning parameters derived from general vision?
% }

% \smallskip

% Existing works \cite{qian2023nuscenes,wu2023nuprompt,sachdeva2023rank2tell,malla2023drama,kim2018textual,kim2019CVPR} attempt to apply general-purpose VLMs in driving for scene description by using off-the-shelf models \cite{li2023blip2,gao2023llamaadapterv2,liu2023llava,li2023mimicit,2023videochat,Maaz2023VideoChatGPT} and just fine-tuning them for several epochs on driving datasets.
% %
% However, due to the lack of large-scale training on tasks like localization, these models have weak spatial performance (please refer to Tab. {\color{red}3} in the main paper).
% %
% For VLMs to be genuinely applicable in autonomous driving scenarios, a preferable method is to train a VLM that is specifically catered towards the needs of driving, \textit{i.e.}, description, localization, memorization, and forecasting.

\bigskip

\textbf{Q5:} \textit{Why does the benchmark of embodied understanding for autonomous driving incorporate data from the general computer vision domains, e.g., Ego4D \cite{grauman2022ego4d}?
}

\smallskip

Ego4D contains diverse egocentric videos gathered worldwide, with driving scenarios as a subset. 
It offers a more varied and realistic breadth of scenarios, covering a more comprehensive range of situations not included in the original training data. 
The dataset enhances the system's adaptability and improves its ability to handle unpredictable real-world scenes.
Furthermore, behaviors occur over a long duration in Ego4D (such as cooking a dish for several minutes), while those in the driving dataset are pretty brief (such as overtaking a car within a few seconds).
The introduction of the Ego4D dataset facilitates a more reasonable evaluation of the temporal capabilities of VLMs in embodied scene understanding.

\section{Related Work}

The related work is provided below due to the page limit in the main paper.

\subsection{Embodied Understanding}

Embodied understanding aims to allow intelligent agents to follow human instructions, interact with open environments, and incorporate common sense into reasoning \cite{grauman2022ego4d, padalkar2023open,majumdar2023we}.
Recently, we have witnessed the success of embodied understanding, especially in robotics \cite{dipalo2023unified,hao2022language,huang2023language,wang2022git}.
RT-1~\cite{brohan2022rt} realizes an end-to-end pipeline from human instructions to robot control signals in an embodiment setting.
PaLM-E~\cite{driess2023palme} encodes images into visual tokens and integrates them with prompts to generate answers or robotics control instructions. 
It focuses on implementing task decomposition from high-order instruction to low-order execution.
%
% It proves that training on a mixture of diverse tasks across robot manipulation and general vision-language tasks enhances the zero-shot generalization capability. 
%
RT-2~\cite{rt22023arxiv} goes further by realizing a closed loop from the input high-order commands and images to the output robot control signals.
% enables the conversion of robotic actions into textual tokens, seamlessly integrating them into the training pipeline of vision-language models. 
% This integration significantly enhances the model's ability to generalize across novel objects and comprehend instructions with diverse semantic nuances.
%
Even though they can achieve a recurrent output of control signals, they can not query past events required for driving scenarios.
EmbodiedGPT~\cite{mu2023embodiedgpt} extracts embodied-task-specific features from planning queries, enabling robots to perceive their environment and make reasoned decisions.
Voltron~\cite{karamcheti2023voltron} builds a new evaluation suite covering five different robot learning problems, serving as a unified platform for comprehensively evaluating visual representations of embodied understanding for robotics.
RT-Trajectory~\cite{gu2023robotic} uses roughly drawn sketches to provide interactive guidance, aiding the model in accomplishing complex control tasks.
Despite the success of these works in embodied understanding, the models mostly confined to indoor scenes have difficulty being directly applied to driving scenarios.
%, which are characterized by large spatial and temporal spans.
%
% ViNT~\cite{shah2023vint} and its preceding works \cite{shah2022robotic,shah2023gnm,shah2021ving,hirose2023exaug} achieve outdoor navigation in the open world, but these works are limited to localization capabilities, which are far from sufficient understanding of the scene.
%\HY{In contrast, our work differentiates from them in that XXXXX.}
% In contrast, 
Differently,
our work offers augmented large-scale spatial localization capabilities and long-horizon temporal modeling capabilities, extending it to outdoor driving scenarios.
%designs a space-aware pre-training strategy and time-aware token selection module to address the driving scenarios-specific issues.

%
% Despite the success of these works in embodied understanding, it is difficult to apply these efforts to driving scenarios. 
%

%
% This is because they are mostly limited to indoor scenarios and still fail to address large-scale spatial localization. 
% %
% And even though they can achieve recurrent output of control signals, they still can't do the querying of past events required for self-driving scenarios.

\subsection{Large Vision-Language Models}

Visual language models typically serve as the core of embodied scene understanding.
BLIP2~\cite{li2023blip2} offers a universal and efficient pre-training strategy, guiding visual language pre-training from ready-made image encoders and frozen language models.
Flamingo~\cite{alayrac2022flamingo} builds a model that can quickly adapt to new tasks with only a few annotated examples, bridging the gap between powerful pre-trained visual and language models.
LLaMA-Adapter V2~\cite{gao2023llamaadapterv2} unlocks more learnable parameters of the language model. It employs an early fusion strategy by only inputting visual tokens into early LLM layers, thus facilitating better visual knowledge integration.
LLaVA~\cite{liu2023llava} first tries to generate multi-modal instruction data using GPT-4 \cite{openai2023gpt}. LLaVA shows impressive multi-model chatting capabilities, sometimes exhibiting GPT-4 behavior on unseen images or instructions.
The models mentioned above achieve good image-based scene understanding tasks but need to support the inquiry of video content.

Otter~\cite{li2023mimicit} proposes a dataset of 2.2 million instruction-response pairs from images and videos. 
Each pair is accompanied by multi-modal information, forming a dialogue context to enhance VLM's perception, reasoning, and planning capabilities.
VideoChat~\cite{2023videochat} expands the learnable parameters of the model to adapt to the understanding of video content and proposes a tutorial dataset containing thousands of videos with detailed descriptions and dialogues.
By merging the video-adaptive visual encoder and language model, Video-ChatGPT~\cite{Maaz2023VideoChatGPT} can understand and generate human-like video dialogues.
However, these methods of extracting a few frame samples from an entire video cannot fully comprehend the content of a long video.
MovieChat~\cite{song2023moviechat} develops a memory mechanism and enables the summarization of an entire movie. 
Nevertheless, it does not support querying events that happened at a past moment.
Moreover, the above methods can only provide a rough inquiry of relative position and cannot obtain accurate 3D locations.

%To sum up, all models are inherently incapable of comprehensively understanding driving scenarios.
%
In contrast, our work differentiates from the prior VLMs by designing a space-aware pre-training strategy and time-aware token selection module to address the driving scenarios-specific issues.

% In addition, their tasks emphasize recurrently outputting control commands rather than queries about past events.

\subsection{Vision-Language Models for Autonomous Driving}

To introduce embodied understanding into driving scenarios, a series of datasets \cite{wu2023nuprompt,xu2020explainable,deruyttere2022talk2car,wu2023referring,li2023_datasetsurvey} have been proposed.
For the use of explaining driving behaviors, some works \cite{qian2023nuscenes,kim2018textual,kim2019CVPR,keysan2023text,malla2023drama,sima2023drivelm,sachdeva2023rank2tell,echterhoff2023driving} provide annotations for scene descriptions, traffic element analysis, high-level instructions, and danger warnings.
HiLM-D~\cite{ding2023hilmd} first uses natural language to simultaneously recognize and explain risk objects, understand the intentions of the ego-vehicle, and provide motion suggestions.
One work~\cite{chen2023drivingwithllms} introduces a quality assessment metric for driving behavior and demonstrates the proficiency of the VLM driver in interpreting driving scenarios, answering questions, and making decisions.
Adapt~\cite{jin2023adapt} provides user-friendly natural language narratives for each vehicle control and action decision step.
DriveGPT4~\cite{xu2023drivegpt4} achieves an interpretable end-to-end autonomous driving system using a language model, which can explain vehicle actions and provide corresponding reasoning.
Lingo-1~\cite{wayve2023lingo1} integrates vision, language, and action to enhance the industry's interpreting, explaining, and training foundational driving models.
Nevertheless, these works are similar in that they describe the entire scene.

Additionally, some approaches leverage language models to enhance traditional autonomous driving tasks.
An attempt~\cite{elhafsi2023semantic} introduces a monitoring framework for semantic anomaly detection in vision-based policies to achieve open-vocabulary object detection.
However, it can not achieve the spatial localization of concern in driving.
%
% LCTGen~\cite{tan2023lctgen} leverages language as source supervision for generating dynamic traffic scenarios.
%
%
MotionLM~\cite{seff2023motionlm} redefines multi-agent motion prediction as a language modeling task.
However, it only predicts the orientation of the objects over continuous time.
LanguageMPC~\cite{sha2023languagempc} develops an algorithm to convert VLM decisions into actionable driving instructions.
GPT-Driver~\cite{mao2023gptdriver} leverages large language models' inherent powerful reasoning capabilities and generalization potential to achieve trajectory prediction.
Essentially, they use the language model as the decision-maker, leveraging the positioning information given by the detection model to predict trajectories. 
The models themselves do not have spatial localization capabilities.
Furthermore, none of the above methods can query past events over a long time series.

Previous endeavors are constrained to providing descriptions of driving scenes. 
% In contrast, 
Instead,
our study 
% systematically 
analyzes the requirements of driving scenarios, articulates the four essential capabilities, and establishes an evaluation benchmark.

%In contrast, our work distinguishes the prior attempts by offering augmented large-scale spatial localization capabilities and long-horizon temporal modeling capabilities.

\section{ELM - Implementation Details}
\label{sec:implementation_details}

% \section{Implementation Details}

\subsection{Space-aware Pre-training}
\label{sec:location_labeling}

% \begin{figure}[t]
% \centering
% \includegraphics[width=0.7\linewidth]{figs/location_labeling_v2.pdf}
% \caption{
% \textbf{Human-check in the loop 
% % annotation workflow 
% for localization labeling.}
% \texttt{Node} 2-3 represent the selection of question templates generated by GPT-4 \cite{openai2023gpt} by humans.
% \texttt{Node} 4 indicates the generation of location query ground truth from the point cloud and image pairs in nuScenes \cite{caesar2019nuscenes}.
% }
% \label{fig:loc_labeling}
% \end{figure}

\smallskip \noindent \textbf{Auto-labeling with human in the loop.}
To enhance the spatial understanding capability of the model, we design a localization labeling pipeline based on nuScenes \cite{caesar2019nuscenes}.
%
% As depicted in \cref{fig:loc_labeling},
As depicted in Fig. {\color{red}3} of the main paper,
the annotation process involves the manual quality check and is divided into four steps.
In \texttt{Node} 1, GPT-4 \cite{openai2023gpt} generates many unique text prompt templates for diversity.
We provide an instance of prompts to generate templates as follows, where $u$ and $v$ are pixel coordinates.
{
\captionsetup{type=table}
\begin{tcolorbox}[colback=gray!10,%gray background
                  colframe=black,% black frame colour
                  width=\linewidth,
                  arc=1mm, auto outer arc,
                  boxrule=0.5pt,
                 ]
\texttt{\textcolor{blue}{Prompt} = "}Question: Generate 20 diverse templates into a list that convey similar meanings to the following Python statement:

{\color{gray}
\texttt{"}Determine the spatial coordinates in 3D corresponding to the 2D pixel at $<\text{c}, \{u\}, \{v\}>$.\texttt{"},

\texttt{"}Find the 3D spatial coordinates corresponding to the 2D pixel at $<\text{c}, \{u\}, \{v\}>$.\texttt{"},

\texttt{"}Provide the 3D scene position of the 2D point $<\text{c}, \{u\}, \{v\}>$.\texttt{"},

\texttt{"}Compute the 3D scene position of the 2D pixel at $<\text{c}, \{u\}, \{v\}>$.\texttt{"}, \textit{etc.}}

Answer: \texttt{"}
\end{tcolorbox}

}

In \texttt{Node} 2-3, a manual approach is employed to sample and select one thousand high-quality templates.
For the ground truth labels of localization, we establish the correspondence between 3D pixels and 3D points using point clouds and camera parameters.
For the nuScenes dataset, we collected point cloud-image pairs at a sampling rate of 2Hz. 
During the sampling process in \texttt{Node} 4, we validate several sampling strategies:
1) random sampling; 2) sampling based on the pixel distance on the image; 3) selecting only foreground points; 4) sampling the farthest points based on spatial distance.
During the algorithm iteration process, the fourth strategy is the most effective.
We set the sampling threshold to 1.5 meters and ultimately select about 200 points from each frame to construct inquiries about spatial locations.
In the end, we combine the templates with the positional labels to obtain a total of 7.4 million QA pairs related to localization.

Besides, cross-dataset description labels are provided to ensure the high generalization of the model in driving scenarios. 
The pipeline is shown in Fig. {\color{red}3} from the main paper.
Firstly, the annotation process needs to filter out poor-quality images, as the noise introduced by these images can affect the VLM's accuracy.
Taking YouTube data as an example, videos are segmented into continuous image frames at a rate of one frame every five seconds. 
We sample videos with resolutions no less than 720p (\textit{e.g.}, 1280×720 for 16:9 videos) and discard the first 90 seconds and the last 30 seconds for most videos to remove the channel introduction at the beginning and the subscription reminder at the end.
A continuous set of 100 frames is compiled and 10 frames are randomly selected from these to be sent to the annotator.
The annotator will simultaneously browse and compare 100 images from 10 different sets, choosing the worst 1 to 3 sets of images based on indicators such as scene richness, lighting, blurriness, and viewpoint. 
All images from this poorer set are labeled for return and sent back into the data pool, while the remaining nine sets of images pass the screening. 
When a video source is marked for return five times, it is added to a blacklist and annotators no longer receive image data from it.

For the filtered images, we give three types of prompts to LLaMA-Adapter V2 \cite{gao2023llamaadapterv2} to generate labels related to scene captions, traffic elements, and driving behaviors.
Specifically, the example prompts for nuScenes \cite{caesar2019nuscenes}, Waymo \cite{sun2020scalability}, and YouTube are as follows.
{
\captionsetup{type=table}
\begin{tcolorbox}[colback=gray!10,%gray background
                  colframe=black,% black frame colour
                  width=\linewidth,
                  arc=1mm, auto outer arc,
                  boxrule=0.5pt,
                 ]
\texttt{\textcolor{blue}{Prompt\_A} = "}Question: Describe the scenario, especially the unusual ones, to propose suggestions. Answer: \texttt{"}

\texttt{\textcolor{blue}{Prompt\_B} = "}Question: Describe the traffic elements in detail, especially focus on traffic signals, cars, pedestrians and anything vital to driving. Answer: \texttt{"}

\texttt{\textcolor{blue}{Prompt\_C} = "}Question: Which lane is the ego-vehicle driving in and how should we drive at the moment. Answer: \texttt{"}
\end{tcolorbox}
}

As for Ego4D \cite{grauman2022ego4d}, its prompt only involves scene description.
{
\captionsetup{type=table}
\begin{tcolorbox}[colback=gray!10,%gray background
                  colframe=black,% black frame color
                  width=\linewidth,
                  arc=1mm, auto outer arc,
                  boxrule=0.5pt,
                 ]
\texttt{\textcolor{blue}{Prompt} = "}Question: Describe the scene, including the objects and the actions. Answer: \texttt{"}
\end{tcolorbox}
}

Similar to the previous round, for a group of continuous QAs, every 10 are grouped together, with 1 randomly selected and given to the annotator. 
The annotator follows certain guidelines, checking elements like overall environment, road components, traffic lights, lane lines, motion status, and behavior decisions, and chooses the worst 1 to 3 QAs from 10 groups based on these criteria.
The images from this set are returned to the VLM for re-annotation with feedback from the annotator, such as "the annotation of a red traffic light is wrong". 
The annotator will quickly browse images that have passed the review and correct obvious minor errors.

\subsection{Time-aware Token Selection}
\label{sec:token_selection}

\begin{figure}[t]
  \centering
  \includegraphics[width=0.7\linewidth]{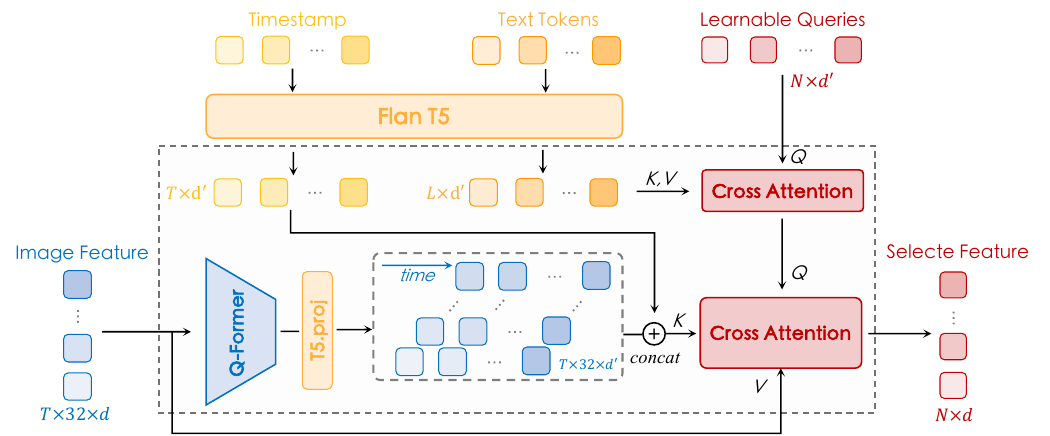}
   \caption{\textbf{Detailed design of the proposed token bank.} 
    This module receives inputs of image features, timestamps, and text tokens. To incorporate the image domain into the text space, we employ the Q-former \cite{li2023blip2}. The process enables the model to perform cross-attention with the text tokens, allowing the model to select the most pertinent tokens from the videos for input to the subsequent language model.}
   \label{fig:token_selection}
\vspace{-0.5cm}
\end{figure}

\cref{fig:token_selection} illustrates the pipeline of our token selection process, and the entire module is referred to as the token bank. 
The input to this module comes from image features $q_v$, timestamps $T_t$, text prompts $T_p$, and learnable queries $q_i$.
The image feature is projected into textual domain features $\hat{q_v} \in \mathbb{R}^{T\times 32\times d'}$ using Q-former with ({\color{red}1}).
The timestamps and input prompts are encoded using FlanT5 \cite{2020t5} to obtain the corresponding tokens.
As referred to as ({\color{red}2}), this module involves two cross-attention mechanisms. 
The learnable queries interact with the text prompts in the first cross-attention to map the variable-length text prompts to a fixed length of $N$ tokens.
The intermediate tokens $q_{\text{mid}}$ are considered to contain the features of the prompts.
In the second cross-attention, $q_{\text{mid}}$ serves as the query, the concatenation of $\hat{q_v}$ and $\text{T5}_{\text{Enc}}(T_{t})$ serves as the key, and $q_v$ serves as the value.
The model selects $N$ tokens representing the corresponding timestamp and image content based on the input prompt requirements during this process.

The process, as mentioned earlier, is referred to as soft selection. 
In addition, we also explore two other alternatives: hard selection and manual selection.
Hard selection refers to the process of selecting the $N$ closest frames and their tokens based on the feature similarity:
\begin{equation}
\centering
    \begin{aligned}
    Q &= \texttt{MHCA}\Big [ q_i, \text{T5}_{\text{Enc}}(T_{p}), \text{T5}_{\text{Enc}}(T_{p})\Big ], \\
    K &= \texttt{concat}\left(\hat{q_v}, \text{T5}_{\text{Enc}}(T_{t})\right), \\
    S &= \texttt{avg\_pool}(\frac{Q\cdot K}{\sqrt{d'}}) \in \mathbb{R}^{T}, \\
    E_{\text{vis}} &= q_{v}[\texttt{argTopN}(S)] \in \mathbb{R}^{N\times 32 \times d}.
    \end{aligned}
\label{eq:hard_selection}
\end{equation}
Hard selection is more deterministic than soft selection as it does not involve probabilistic or weighted selection. 
It also comes with a higher risk of performance degradation when the selection is incorrect. 
As the name suggests, manual selection involves human experts selecting the frames that best represent the desired timestamps. 
In theory, it represents an upper bound for hard selection.
Experiment results in Sec. {\color{red}4.2} show that soft selection tends to achieve the best performance.

\subsection{Planning}

Through the newly formulated tasks of embodied understanding, we have achieved a comprehensive understanding of the driving scenes. 
Further on, autonomous driving ultimately requires guidance on how to drive. 
Therefore, we further extend the model to accomplish the downstream task of planning.
% , which involves taking input instructions and generating the future trajectory points for the ego-vehicle within the next 3 seconds.
%
For this task, the inputs are a sequence of 3 images taken at 0.5-second intervals, direction instructions (\texttt{turn\_left}, \texttt{turn\_right}, and \texttt{keep\_forward}), and current velocity $s$, while the output consists of 6 trajectory points at 0.5-second intervals.
The design of the tokenizer is equivalent to that in Sec. {\color{red}3.2}.

To ensure the diversity of questions for the planning task, we also use GPT-4 \cite{openai2023gpt} to generate as diverse a set of question templates as possible. The example prompt is provided below.
{
\captionsetup{type=table}
\begin{tcolorbox}[colback=gray!10,%gray background
                  colframe=black,% black frame color
                  width=\linewidth,
                  arc=1mm, auto outer arc,
                  boxrule=0.5pt,
                 ]
\texttt{\textcolor{blue}{Prompt} = "}Question: Generate 20 diverse templates into a list that convey similar meanings to the following Python statement: 

{\color{gray}
\texttt{"}The ego car is moving $\{\text{direction}\}$ at a speed of $\{s\}$. Predict six trajectory points in the future.\texttt{"},

\texttt{"}Determine the trajectory of the ego car, moving \{direction\}, with a speed of $\{s\}$ for the next 6 points.\texttt{"},

\texttt{"}Predict the future trajectory of the ego car, traveling \{direction\} at a speed of $\{s\}$, for the next 6 points.\texttt{"},

\texttt{"}Calculate the trajectory points for the ego car, which is moving \{direction\} at a speed of $\{s\}$, for the next six instances.\texttt{"}, \textit{etc.}}

Answer: \texttt{"}
\end{tcolorbox}

}

\section{ELM - Benchmark}

\subsection{Fine-tuning Datasets}
Our fine-tuning datasets correspond to all tasks and are built upon many popular datasets, including nuScenes~\cite{caesar2019nuscenes} and Ego4D~\cite{grauman2022ego4d}.
The data sources and label formats for each task are as follows:
\begin{itemize}
    \item \textit{Surrounding Narration:} It is based on DriveLM \cite{sima2023drivelm}, which is annotated with QAs regarding the object categories, presence, and occlusion in the current scene in the nuScenes dataset.
    \item \textit{Traffic Sign Inquiry:} The task is built upon Openlane-V2 \cite{wang2023openlanev2}, which is annotated with road signs and traffic lights corresponding to each lane in the nuScenes dataset. 
    Similar to the generation in Sec. {\color{red}3.2}, we use the templates generated by GPT-4 \cite{openai2023gpt} combined with traffic sign labels to form QA pairs.
{
\captionsetup{type=table}
\begin{tcolorbox}[colback=gray!10,%gray background
                  colframe=black,% black frame color
                  width=\textwidth,
                  arc=1mm, auto outer arc,
                  boxrule=0.5pt,
                 ]
\texttt{\textcolor{blue}{Template Examples:}}

{\color{gray}
\texttt{"}Did the ego vehicle encounter any traffic sign earlier?\texttt{"},

\texttt{"}Has the driver observed any traffic sign previously?\texttt{"},

\texttt{"}Did the car detect any traffic sign before?\texttt{"}, \textit{etc.}
}
\end{tcolorbox}
}    
    \item \textit{Action \& Decision:} QAs related to objects' motion and the ego vehicle's planning instructions are derived from DriveLM.
    \item \textit{Box Detection:} Based on the methodology described in Sec. {\color{red}3.2}, we generate QAs regarding the object positions and categories from the nuScenes labels.
{
\captionsetup{type=table}
\begin{tcolorbox}[colback=gray!10,%gray background
                  colframe=black,% black frame colour
                  width=\textwidth,
                  arc=1mm, auto outer arc,
                  boxrule=0.5pt,
                 ]
\texttt{\textcolor{blue}{Template Examples:}} 

{\color{gray}
\texttt{"}What are the 3D coordinates for the 2D pixel at $<\text{c}, \{u\}, \{v\}>$.\texttt{"}, \textit{etc.}
}
\end{tcolorbox}
}  

    \item \textit{Tracking:} Leveraging the tracking labels in nuScenes, we incorporate the past timestamps $t$ into the Box Detection to formulate this task.
{
\captionsetup{type=table}
\begin{tcolorbox}[colback=gray!10,%gray background
                  colframe=black,% black frame colour
                  width=\textwidth,
                  arc=1mm, auto outer arc,
                  boxrule=0.5pt,
                 ]
\texttt{\textcolor{blue}{Template Examples:}}

{\color{gray}
\texttt{"}Calculate the 3D position of the 2D pixel at $<\text{c}, \{u\}, \{v\}>$ $\{t\}$ seconds ago.\texttt{"}, \textit{etc.}
}
\end{tcolorbox}
}  
    \item \textit{Box Prediction:} Like the previous statement, we use future timestamps $t$ this time.
{
\captionsetup{type=table}
\begin{tcolorbox}[colback=gray!10,%gray background
                  colframe=black,% black frame colour
                  width=\textwidth,
                  arc=1mm, auto outer arc,
                  boxrule=0.5pt,
                 ]
\texttt{\textcolor{blue}{Template Examples:}}

{\color{gray}
\texttt{"}Determine the coordinates in 3D space to the 2D pixel at $<\text{c}, \{u\}, \{v\}>$ $\{t\}$ seconds later.\texttt{"}, \textit{etc.}
}
\end{tcolorbox}
}    
    \item \textit{Egocentric Narration:} We use the timestamps provided by Ego4D \cite{grauman2022ego4d} for each narration to extract the corresponding image from the video. These images and narrations serve as the data for the task. 
{
\captionsetup{type=table}
\begin{tcolorbox}[colback=gray!10,%gray background
                  colframe=black,% black frame color
                  width=\textwidth,
                  arc=1mm, auto outer arc,
                  boxrule=0.5pt,
                 ]
\texttt{\textcolor{blue}{Template Examples:}}

{\color{gray}
\texttt{"}Give a caption for this image.\texttt{"},

\texttt{"}Describe the scene.\texttt{"}, \textit{etc.}
}
\end{tcolorbox}
}      
    \item \textit{Moment Recap:} Timestamps are chosen from a 60-second video, and we ensure they are at least 20 seconds apart from the current moment. The corresponding images and narrations of these timestamps $t$ are subsequently utilized as the data.
{
\captionsetup{type=table}
\begin{tcolorbox}[colback=gray!10,%gray background
                  colframe=black,% black frame colour
                  width=\textwidth,
                  arc=1mm, auto outer arc,
                  boxrule=0.5pt,
                 ]
\texttt{\textcolor{blue}{Template Examples:}}

{\color{gray}
\texttt{"}What took place $\{t\}$ seconds in history?\texttt{"},

\texttt{"}Can you recount the historical event that took place $\{t\}$ seconds back?\texttt{"}, \textit{etc.}
}
\end{tcolorbox}
}  
    \item \textit{Event Query:} Three consecutive frames are selected from a 60-second video. The narrations of the first and third frames are utilized as the question, while the narration of the middle frame is employed as the answer.
{
\captionsetup{type=table}
\begin{tcolorbox}[colback=gray!10,%gray background
                  colframe=black,% black frame colour
                  width=\textwidth,
                  arc=1mm, auto outer arc,
                  boxrule=0.5pt,
                 ]
\texttt{\textcolor{blue}{Template Examples:}}

{\color{gray}
\texttt{"}Tell me about the events that took place between \{Event\_A\} and \{ Event\_B\}.\texttt{"},

\texttt{"}Can you describe what occurred between \{Event\_A\} and \{ Event\_B\}?\texttt{"}, \textit{etc.}
}
\end{tcolorbox}
}  
    \item \textit{Activity Prediction:} The narration related to a future timestamp is extracted as the data.
{
\captionsetup{type=table}
\begin{tcolorbox}[colback=gray!10,%gray background
                  colframe=black,% black frame colour
                  width=\textwidth,
                  arc=1mm, auto outer arc,
                  boxrule=0.5pt,
                 ]
\texttt{\textcolor{blue}{Template Examples:}}

{\color{gray}
\texttt{"}What event will occur in the next $\{t\}$ seconds in the future?\texttt{"}, \textit{etc.}

}
\end{tcolorbox}
}  
\end{itemize}

\subsection{Metrics}

% \smallskip 
\noindent \textbf{Language Metrics.}

\textbf{BLEU} (Bilingual Evaluation Understudy)~\cite{papineni2002bleu} measures the similarity between a generated text and the reference texts. 
It compares $n$-grams (a continuous group of $n$ words) in the generated text to those in the reference texts, with higher precision indicating a better match. 
The BLEU score exhibits insensitivity to semantic nuances and variations in word order.

\textbf{ROUGE\_L} (Recall-Oriented Understudy for Gisting Evaluation)~\cite{lin2004rouge} calculates scores with the longest common subsequence of the model outputs and the reference answers. 
Like the BLEU metric, ROUGE\_L is used to assess the level of matching between generated results and standard references, with the critical difference being that ROUGE\_L is based on recall. 
It provides higher scores for matching longer sequences, thus awarding higher scores to summaries that contain more shared content with the source text.

% \bheading{METEOR}~\cite{lavie-agarwal-2007-meteor} takes into account precision, recall, stemming, synonymy, stemming, and word order. 
% % 
% It establishes alignment between model outputs and references, computes the 1-gram matching between them, and then applies penalties based on chunk blocks, providing a more nuanced evaluation.
% 
% METEOR provides a more nuanced evaluation, capturing not only lexical similarities but also structural and syntactic aspects. 

\textbf{CIDEr} (Consensus-based Image Description Evaluation)~\cite{vedantam2015cider} combines elements from BLEU and vector space models. 
It quantifies the similarity between human-written and machine-generated descriptions using $n$-grams, where the $n$-grams are weighted according to their frequency of occurrence in the human-written descriptions.
Therefore, it captures precision and recall and evaluates the consensus among multiple human references.

\textbf{$A_{\text{GPT}}$ and $S_{\text{GPT4V}}$} refer to utilizing the powerful language and logical reasoning capabilities of GPT4 \cite{openai2023gpt} or GPT4V \cite{openai2023gpt4v} to evaluate text pairs or image-text pairs. 
Traditional metrics primarily evaluate performance at the word level and may not capture subtle semantic differences. 
The powerful reasoning capabilities of ChatGPT can be used to assess the quality of predictions and yield more rational scores. ChatGPT is prompted to assign numeric scores between 0 and 100, with higher scores indicating higher predictive accuracy. Details for the evaluation prompt are as follows.

{
\captionsetup{type=table}
\begin{tcolorbox}[colback=gray!10,%gray background
                  colframe=black,% black frame colour
                  width=\textwidth,
                  arc=1mm, auto outer arc,
                  boxrule=0.5pt,
                 ]
\texttt{\textcolor{blue}{Prompt} = }
\texttt{"}Rate my answer based on the ground truth answer from 0 to 100, with higher
scores indicating that the answer is closer to the ground truth, and you should be accurate to single digits like 62, 78, 41, etc. This is the correct answer: \{{\color{blue}GT}\}. This is my answer: \{{\color{blue}Pred}\}.\texttt{"}
\end{tcolorbox}
}

{
\captionsetup{type=table}
\begin{tcolorbox}[colback=gray!10,%gray background
                  colframe=black,% black frame colour
                  width=\textwidth,
                  arc=1mm, auto outer arc,
                  boxrule=0.5pt,
                 ]
\texttt{\textcolor{blue}{Prompt} = }
\texttt{"}Please score from 0 to 100 based on whether the given image and text correctly match. The higher the score, the more accurate and comprehensive the text description of the image is.
You should give single digits like 62, 78, 41, etc. This is the image: \{{\color{blue}Img}\}. This is my answer: \{{\color{blue}Pred}\}.\texttt{"}
\end{tcolorbox}
}

\textbf{$D_{\text{n-gram}}$} is used to straightforwardly measure the text diversity of the corpus. 
Specifically, each sentence in the corpus needs to be tokenized into individual words. 
The ratio of the number of unique tokens to the number of all tokens, considering repetitions, serves as this metric.

\textbf{Discussion.} CIDEr is considered superior to BLEU and ROUGE\_L \cite{aafaq2019video}.
As for the BLEU metric, a concise reply might lower the score since shortening the reply reduces the chance of a match for BLEU. 
As CIDEr focuses on semantic consistency, it is less likely to be influenced by the length of the reply sequence.
In addition, BLEU and ROUGE\_L may overlook some important information in the text, such as semantic meanings and critical contextual content. 
Alternatively, CIDEr assigns greater weight to $n$-grams frequently appearing in human-written descriptions, thus accurately capturing critical information in the text. 
If an $n$-gram frequently appears in multiple descriptions, it is likely crucial for describing the image. 
% Therefore, CIDEr tends to reward generated descriptions that contain these $n$-grams, capturing the consistency with reference descriptions more effectively. 
Additionally, CIDEr introduces TF-IDF (Term Frequency-Inverse Document Frequency) to measure the significance of $n$-grams, thus accurately capturing essential information in descriptions.
The GPT-related metrics are costly and have a slow evaluation speed, making them unsuitable for large-scale evaluation.
Therefore, we use CIDEr as the primary metric.

% % 
% However, CIDEr has limitations, including the dependency on a large-scale image description corpus for TF-IDF weight calculation and the inability to account for the grammar and structure of sentences.

% \bheading{SPICE}~\cite{anderson2016spice} first parses the text into a syntactic dependency tree using Probabilistic Context-Free Grammar~\cite{jelinek1992basic}, then maps the dependency tree into a scene graph in a rule-based manner.
% %
% The scene graph describes the objects, attributes, and their relationship in the original text, and the SPICE score is computed as the F-score of the generated scene graphs from prediction and ground truth.

% \bheading{GPT Score} is a metric provided by ChatGPT. Traditional metrics mainly assess word-level performance and may not capture semantic nuances, potentially yielding unexpected evaluation outcomes. Leveraging ChatGPT's robust reasoning capabilities, we employ it to gauge prediction quality and derive a more rational score. ChatGPT is prompted to assign a numerical score between 0 and 100, with higher scores indicative of enhanced prediction accuracy. The detailed prompt for GPT score evaluation is shown in Table~\ref{box:gpt_score}.

% \subsection{Diversity of Pre-training Dataset}
% \label{sec:diversity_data}

\smallskip 
\noindent 
\textbf{Localization Metrics.}

For localization tasks, the language mentioned above evaluation metrics are no longer applicable. 
The reason lies in the semantic proximity of numbers, making it challenging to reflect physical distance discrepancies via language evaluation metrics. 
For instance, 0.12 and 12 might look similar, but there is a significant difference in their numeric values.
%
% It is imperative, therefore, to establish an evaluation metric based on the numbers themselves. 
The question-and-answer manner produces no confidence scores, so the traditional 3D detection evaluation metric mAP cannot be utilized directly. 
So in ({\color{red}3}), we introduced a new evaluation metric, Pr@k, to assess localization tasks, taking into account both the accuracy of classification and the disparity between predicted coordinates and ground truths.

\section{Experiments}
\label{sec:more_experiment}

\subsection{Protocols} 
\label{sec:fine_tune_dataset}

% , DriveLM~\cite{sima2023drivelm}, OpenLane-V2~\cite{wang2023openlanev2} 
% %
% To illustrate, for Box Detection, we utilize the available annotation data from nuScenes to populate location information for objects in the scene. This is done by applying a rule-based approach within our predefined interrogation templates.
% %
% The expected answer consists of three precise 3D coordinates, followed by the object's category name. 
% %
% Similarly, our fine-tune dataset for Traffic Sign Inquiry is constructed in a similar fashion using OpenLane-V2. 
% %
% The datasets for Perception and Prediction \& Planning tasks are directly imported from DriveLM. 
% % 
% The Ego4D dataset provides a caption for each frame describing the camera wearer's motion. 
% %
% We leverage this information to formulate three types of problems: 
% \uppercase\expandafter{\romannumeral1}. Identifying the action that transpired t seconds ago; 
% \uppercase\expandafter{\romannumeral2}. Determining the action that arose between action A and action B; 
% \uppercase\expandafter{\romannumeral3}. Predicting the most probable action at the t-th second in the future. 
% As the model is provided with up to 20 frames of video spanning between 30s-60s as the visual input, it necessitates robust memory and information extraction capabilities to better solve the challenges.

% \noindent \textbf{Training.} 
%
ELM is first pre-trained on the open-world data corpus and then fine-tuned on diverse tasks.
During the fine-tuning stage, our model and all the evaluated VLMs \cite{li2023blip2,alayrac2022flamingo,gao2023llamaadapterv2,liu2023llava,li2023mimicit,2023videochat,Maaz2023VideoChatGPT} are trained in three separate groups: one group for tasks on Ego4D, another group for tasks related to localization, and a third group for other tasks.
Although we employed group training in our implementation, the language models can potentially train all tasks simultaneously.
In addition, all models are trained until saturation (6 epochs). 
For single-frame models like LLaMA-Adapter V2 \cite{gao2023llamaadapterv2} and LLaVA \cite{liu2023llava}, we employ frame concatenation to create their corresponding multi-frame versions.

As for the training settings, we follow most of the basic settings as in BLIP2-flant5 \cite{li2023blip2}, with a batch size of 8, a learning rate of 1e-5, and AdamW \cite{loshchilov2017decoupled} optimizer with a weight decay of $1\times10^{-2}$.
The warm-up learning rate is $1\times10^{-8}$, with warm-up steps of 1000. 
The input images are resized to $364\times 364$, and the maximum length of prompt tokens is 320.
The number of tokens for each frame in Slot Attention \cite{locatello2020object} is 32, while the number of tokens for Q-Former \cite{li2023blip2} is also 32. 
The token bank is initialized with 128 learnable queries.
Experiments are conducted with 16 NVIDIA Tesla A100 GPUs.

\begin{figure}[t]
  \centering
  \includegraphics[width=0.5\linewidth]{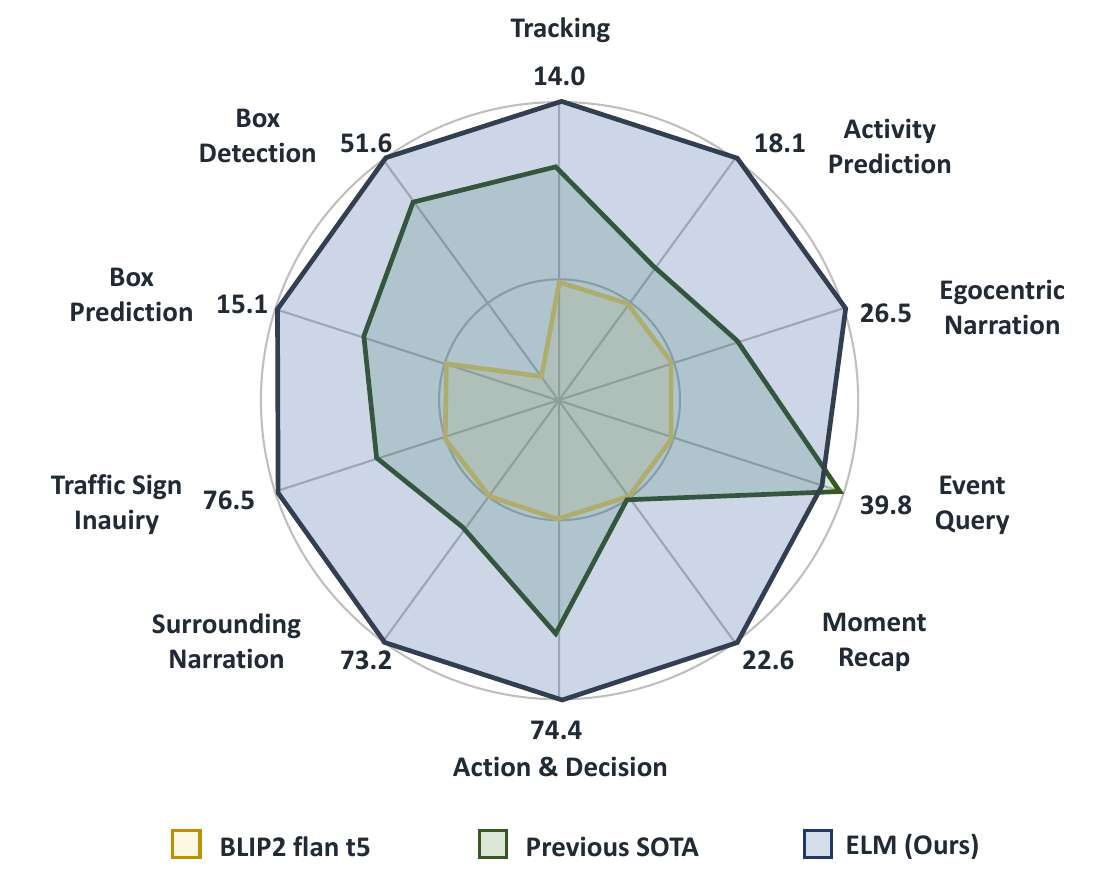}
   \caption{
   \textbf{Qualitative comparisons among BLIP2-flant5 \cite{li2023blip2}, Previous SOTA, and ELM.}
    The metrics for BLIP2-flant5 and ELM are reported, with the previous SOTA as the baseline for each task. It is important to note that the scale of each axis in the radar chart has been adjusted based on the performance of BLIP2-flant5 and ours. The length of the axis is linearly bound to the value.
   }
   \label{fig:radar}
\end{figure}

\begin{table}[t]
\centering
\footnotesize
% \tablestyle{2pt}{1.05}
\setlength{\tabcolsep}{3mm}{
\scalebox{0.73}{
\begin{tabular}{l|cccc|cccc}  
\toprule
\multirow{2}{*}{Method} & \multicolumn{4}{c|}{L2 (m) $\downarrow$} & \multicolumn{4}{c}{Collision (\%)$\downarrow$} \\ 
& 1s & 2s & 3s & \baseline{Avg} & 1s & 2s & 3s & \baseline{Avg}  \\ \midrule
NMP \cite{zeng2019end} & - & - & 2.31 & \baseline{-} & - & -& 1.92 & \baseline{-} \\ 
SA-NMP \cite{zeng2019end}& - & - & 2.05 & \baseline{-} & - & -& 1.59 & \baseline{-} \\ 
FF \cite{hu2021safe} & 0.55 & 1.20 & 2.54 & \baseline{1.43} & 0.06 & 0.17& 1.07 & \baseline{0.43} \\ 
EO \cite{khurana2022differentiable} & 0.67 & 1.36 & 2.78 & \baseline{1.60} & \textbf{0.04} & \textbf{0.09} & 0.88 & \baseline{\textbf{0.33}} \\ 
ST-P3 \cite{hu2022stp3} & 1.33 & 2.11 & 2.90 & \baseline{2.11} & 0.23 & 0.62& 1.27 & \baseline{0.71} \\ 
UniAD \cite{hu2023_uniad} & 0.48 & \textbf{0.96} & \textbf{1.65} & \baseline{\textbf{1.03}} & 0.05 & 0.17& \textbf{0.71} & \baseline{0.35} \\ 
UniAD$^{\dagger}$ \cite{hu2023_uniad} & - & - & - & \baseline{1.80} & - & -& - & \baseline{2.62}  \\\midrule
\textbf{ELM} & \textbf{0.34} & 1.23 & 2.57 & \baseline{1.38} & 0.12 & 0.50& 2.36 & \baseline{0.99} \\ \bottomrule
\end{tabular}
}
}
\vspace{5pt}
\caption{
\textbf{A comparison of ELM to state-of-the-art end-to-end planning approaches.}
Our model achieved comparable results to FF \cite{hu2021safe}, EO \cite{khurana2022differentiable}, ST-P3 \cite{hu2022stp3}, and UniAD \cite{hu2023_uniad}, \textit{etc}.
ELM yields a significant lead on the L2 metric at 1 second.
$\dagger$ : The version of UniAD that only uses the \texttt{camera\_front} images.
\texttt{Avg}: average.
}
\label{tab:comparison_planning}
\underfigtab
\end{table}

\subsection{More Comparisons to State-of-the-arts}

\smallskip \noindent \textbf{Comparison to previous approaches.}
\cref{fig:radar} presents a more intuitive comparison of ELM to other VLMs through a radar chart.
The previous SOTA refers to the results of the best method compared in each task.
We highlight the main metrics corresponding to the outermost circle of the chart.
The maximum and minimum values of the coordinate axis are adjusted while keeping the proportion of the axis constant.
As the chart shows, our model significantly surpasses the baseline in all aspects and notably leads the state-of-the-art in nine tasks.
Our model performs slightly less well on the Event Query, as the Flant5 \cite{2020t5} that ELM is based on has limitations when processing long sequence input text.

\smallskip \noindent \textbf{Quantitative results on planning.}
\cref{tab:comparison_planning} describes a comparison of the ELM with end-to-end driving models on planning.
The results indicate that our model can achieve comparable performance with end-to-end autonomous driving approaches, especially on the L2 metrics.
ELM outperforms models like FF \cite{hu2021safe}, EO \cite{khurana2022differentiable}, STP-3 \cite{hu2022stp3}, \textit{etc.} in terms of accuracy in trajectory prediction.
Our model has the most minor L2 error at 1 second, demonstrating its excellent localization capabilities after space-aware pre-training.

\begin{figure}[t]
\centering
\includegraphics[width=0.7\linewidth]{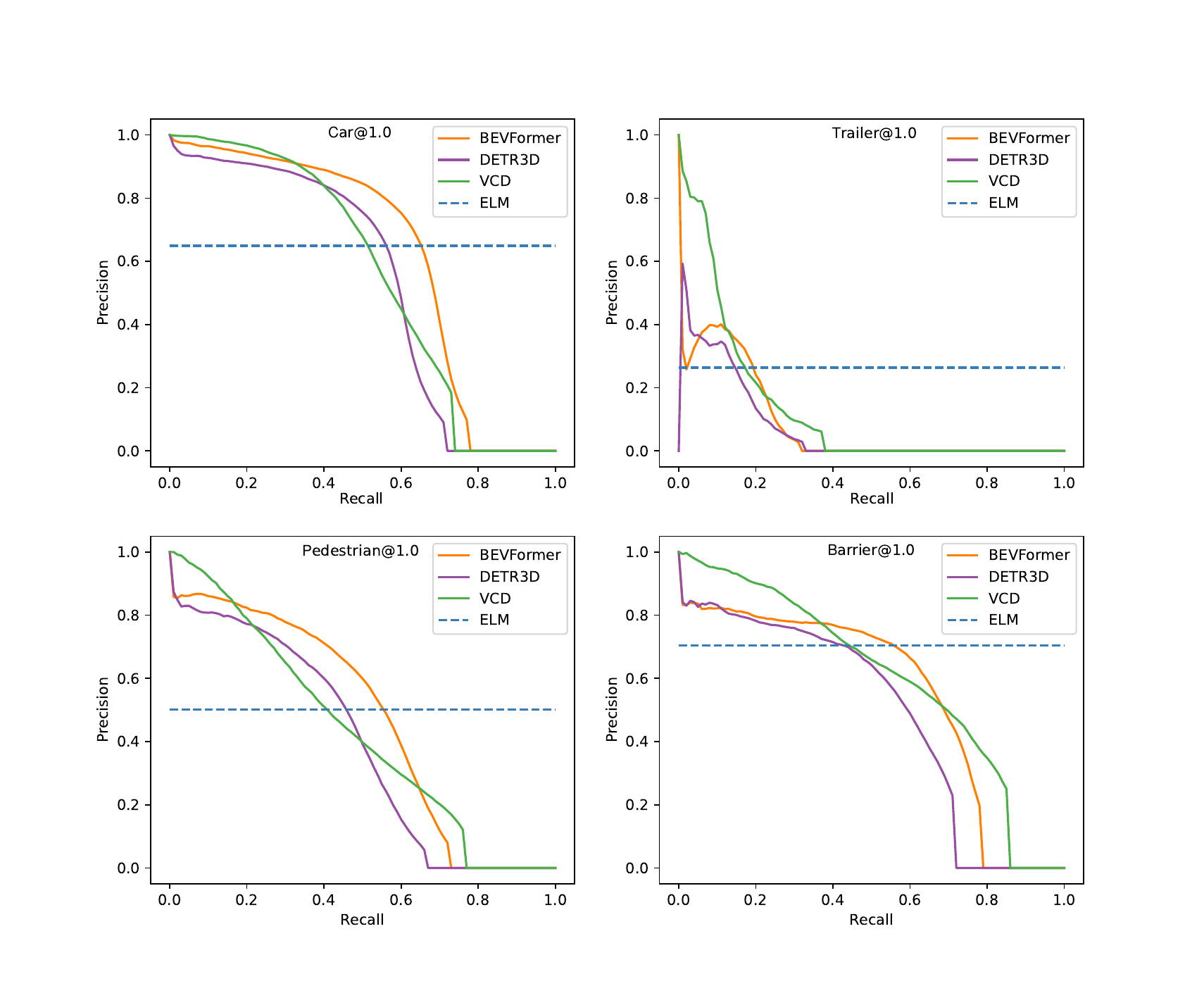}
\caption{
\textbf{Comparison to 3D detection models.}
We provide the precision-recall curves of BEVFormer \cite{li2022bevformer}, DETR3D \cite{wang2022detr3d}, and VCD \cite{huang2023leveraging} under different categories.
Since ELM does not have a recall metric, we use a dashed line to represent the precision of ELM, pointing out where ELM surpasses the accuracy of traditional models at specific recall values.}
\label{fig:pr_chart}
\end{figure}

% \subsection{Out-of-distribution Evaluation}

% table and one figure

\subsection{More Ablations}

\smallskip \noindent \textbf{Ablation on space-aware pre-training.}
To further validate the role of our pre-training dataset in enhancing the localization capability of other VLMs, we supplement experiments on BLIP2-opt \cite{li2023blip2} and Flamingo \cite{alayrac2022flamingo} in \cref{tab:space_pre_train}.
Both models are pre-trained using our localization data and then fine-tuned on the Box Detection task. 
We report the models' Pr@1 and Pr@2 metrics for this task. 
To provide a more intuitive comparison of the pre-training benefits of localization, we also report the non-category-specific Pr@1 and Pr@2 metrics.
The experimental results show that introducing space-aware pre-training can help Flamingo achieve significant improvement in localization-related tasks with \textbf{+8.8\%} and \textbf{+11.2\%} on Pr@1 metrics.

\smallskip \noindent \textbf{Ablation on time-aware token selection.}
Our proposed token selection module has the potential to help other VLMs enhance their performance in tasks related to long-time series.
Given that the textual encoding of timestamps requires the language model's encoder, we report the results of Flamingo \cite{alayrac2022flamingo} in \cref{tab:time_token_selection} only.
It can be observed that the model's performance on Event Query increases by \textbf{+25.3\%}, indicating that our frame can be applied to other encoder-decoder structured VLMs and enhance their memorization and forecasting abilities.

% \begin{table}[t]
% \centering
% \footnotesize
% \setlength{\tabcolsep}{2mm}{
% \scalebox{0.63}{
% \begin{tabular}{l|cc|c}
% \toprule
%  Method & $A_{\text{GPT}}$ & $S_{\text{GPT4V}}$ & $D_{\text{n-gram}}$  \\ \midrule
%  Baseline & 54.3 & 34.4 & 14.8 \\
%  + Filtering & 68.3 & 49.5 & 21.2 \\
%  \baseline{+ Verification} & \baseline{\textbf{84.4}} & \baseline{\textbf{66.9}} & \baseline{\textbf{26.7}}  \\ \midrule
%  \textit{Manual Labeling} & \textit{100} & \textit{64.3} & \textit{23.3} \\
% \bottomrule
% \end{tabular}
% }
% }
% \vspace{+5pt}
% \caption{
% \textbf{Ablations of auto labeling on label quality and diversity.}
% \texttt{Baseline}: LLaMA-Ada., \texttt{$A_{\text{GPT}}$}: 
% accuracy between auto and manually annotated text evaluated by GPT, \texttt{$S_{\text{GPT4V}}$}: rationality score in image-text matching evaluated by GPT4V, \texttt{$D_{\text{n-gram}}$}: 
% diversity evaluated by distinct n-gram ratio of phrases.
% Our choice is marked in \colorbox{baselinecolor}{gray}.
% }
% \label{tab:auto_label}
% \underfigtab
% \end{table}

\begin{table}[t]
\centering
\footnotesize
\setlength{\tabcolsep}{3mm}{
\scalebox{0.63}{
\begin{tabular}{l|c|cc|cc}
\toprule
\multirow{2}{*}{Method}       & \multirow{2}{*}{Pre-training} & \multicolumn{2}{c|}{BD w Category} & \multicolumn{2}{c}{BD w/o Category} \\   
&                           & \baseline{Pr@1}       & Pr@2       & \baseline{Pr@1}        & Pr@2       \\ \midrule
\multirow{2}{*}{BLIP2-opt \cite{li2023blip2}}        & w/o&        \baseline{0.1}    &0.2 & \baseline{5.4} & 8.1\\   
& w  &     \baseline{\textbf{19.4}}       &\textbf{39.5} &\baseline{\textbf{29.3}}  &\textbf{60.9} \\ \midrule
\multirow{2}{*}{Flamingo \cite{alayrac2022flamingo}}        & w/o&   \baseline{19.3}         &31.6 & \baseline{31.5} & 51.0\\   
& w  &       \baseline{\textbf{28.1}}     & \textbf{38.1} & \baseline{\textbf{42.7}} & \textbf{60.1}\\ 
% \midrule
% \multirow{2}{*}{BLIP2-flant5 \cite{li2023blip2}} & w/o&  \baseline{5.1}          &  10.5 &\baseline{8.6}  & 15.3\\   
% & w  &      \baseline{\textbf{51.6}}      &  \textbf{56.9} & \baseline{79.9} & 88.9\\ 
\bottomrule
\end{tabular}
}
}
\vspace{5pt}
\caption{
\textbf{Ablation on space-aware pre-training.}
We pre-train BLIP2-opt \cite{li2023blip2} and Flamingo \cite{alayrac2022flamingo} on the localization data and then fine-tune them on the Box Detection task, respectively. The Pr@1 and Pr@2 metrics on both with category and without category evaluations are reported.
\texttt{BD}: Box Detection.
}
\label{tab:space_pre_train}
\underfigtab
\end{table}

\begin{table}[t]
\centering
\footnotesize
\setlength{\tabcolsep}{2mm}{
\scalebox{0.63}{
\begin{tabular}{l|c|ccc|ccc|ccc}
\toprule
\multirow{2}{*}{Method}       & \multirow{2}{*}{Token Selection} & \multicolumn{3}{c|}{Moment Recap} & \multicolumn{3}{c|}{Event Query} & \multicolumn{3}{c}{Activity Prediction} \\   
&                           & \baseline{CIDEr}    & ROUGE\_L & BLEU     & \baseline{CIDEr}    & ROUGE\_L & BLEU& \baseline{CIDEr}    & ROUGE\_L & BLEU     \\ \midrule
\multirow{2}{*}{Flamingo \cite{alayrac2022flamingo}}        & w/o&        \baseline{11.0}    &  30.1
    &  12.4   & \baseline{15.5} &  32.5  & 14.3 &  \baseline{9.0} &  29.1  &  11.1\\   
& w  &      \baseline{\textbf{18.2}}    &  \textbf{33.9}
    &   \textbf{16.7}  & \baseline{\textbf{40.8}} & \textbf{45.9}   & \textbf{34.8} &  \baseline{\textbf{14.8}} &  \textbf{32.3}  & \textbf{15.5} \\   
\bottomrule
\end{tabular}
}
}
\vspace{5pt}
\caption{
\textbf{Ablation on time-aware token selection.}
Flamingo \cite{alayrac2022flamingo} is used to validate the enhancement effect of our proposed time-aware token selection on Moment Recap, Event Query, and Activity Prediction tasks.
}
\label{tab:time_token_selection}
\underfigtab
\end{table}

% \begin{figure}[t]
% \centering
% \begin{minipage}[b]{\linewidth}
% \includegraphics[width=1\linewidth]{figs/supp_elm_vis_youtube.pdf}
% \subcaption{Youtube}\vspace{4pt}
% \end{minipage}

% \begin{minipage}[b]{\linewidth}
% \includegraphics[width=1\linewidth]{figs/supp_elm_vis_nuscenes.pdf}
% \subcaption{NuScenes}\vspace{4pt}
% \end{minipage}
% \caption{visualization of generated data}
% \end{figure}

\smallskip \noindent \textbf{Comparison to 3D detection approaches.}
In Sec. {\color{red}4.1}, we compare ELM with traditional 3D detection models under the Pr@1 metric. 
For a further comparison, \cref{fig:pr_chart} shows the variance in precision at different levels of recall for BEVFormer \cite{li2022bevformer}, DETR3D \cite{wang2022detr3d}, and VCD \cite{huang2023leveraging}.
Given that ELM does not have a recall metric, we use a dashed line to represent the precision of ELM under this category.
The chart shows that ELM surpasses all 3D detection models for the Car category when recall is more significant than 0.7. 
For the Pedestrian and Barrier category, ELM is the best when recall is more significant than 0.6.
These results demonstrate that our model can achieve comparable results to traditional detection models.

\section{Qualitative Results}
\label{sec:more_visualization}

\subsection{Open World Data Corpus}

\begin{figure}[H]
\centering
\vspace{1.5cm}
\includegraphics[width=\linewidth]{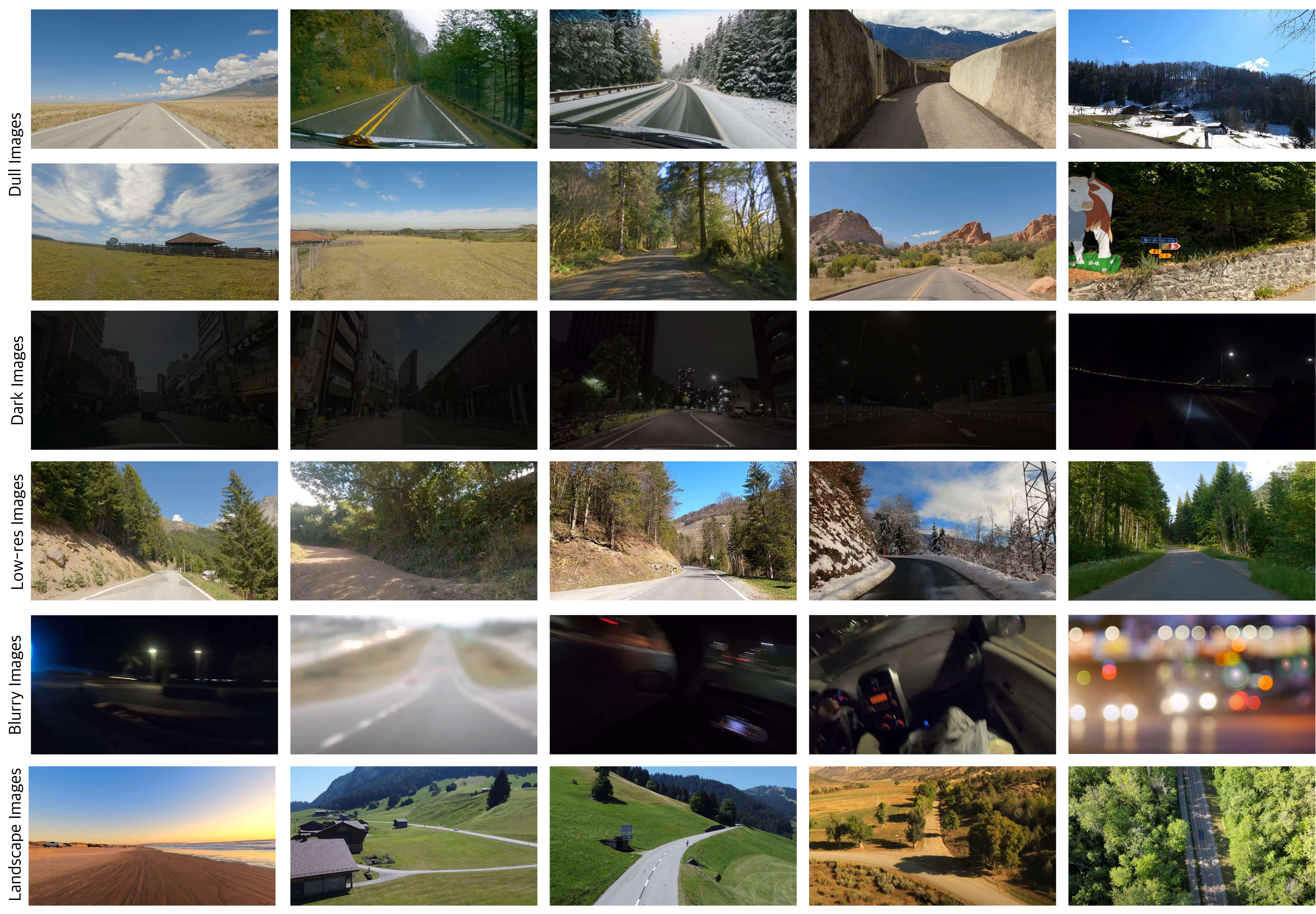}
\caption{
\textbf{Low-quality images to be disposed of.}
The figure shows several types of images that need to be filtered out.
\texttt{res}: resolution.
}
\vspace{-1cm}
\label{fig:filtered_img}
\end{figure}

\begin{figure}[H]
\centering
\includegraphics[width=\linewidth]{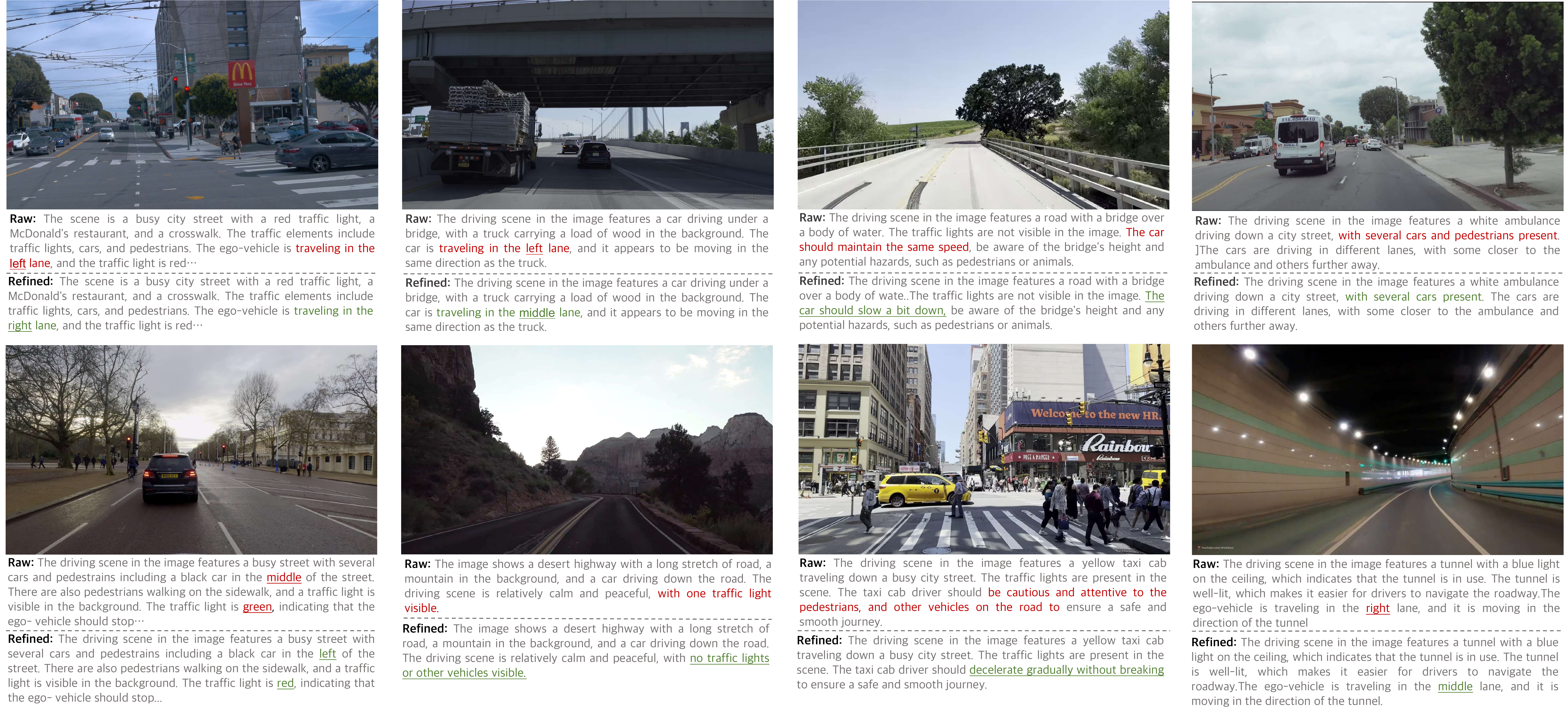}
\caption{
\textbf{Examples of human quality check and manual revision for description annotation.}
}
\label{fig:human_refine}
\end{figure}

\smallskip \noindent \textbf{Annotation with human quality check.}
Ensuring the quality of pre-training data is essential. 
For this reason, we use manual quality checks to ensure that the processes of image data collection and label generation are high quality.
\cref{fig:filtered_img} shows the examples of images discarded due to poor quality through our manual selection process, which occurs in \texttt{node} 2-3 in Sec. {\color{red}3.2}. 
\cref{fig:human_refine} shows the manual revision of labels generated by LLaMA-Adapter V2 \cite{gao2023llamaadapterv2}, and this process takes place in \texttt{node} 7-8 in Sec. {\color{red}3.2}.

\begin{figure}[H]
\centering
\vspace{1.5cm}
\includegraphics[width=0.99\linewidth]{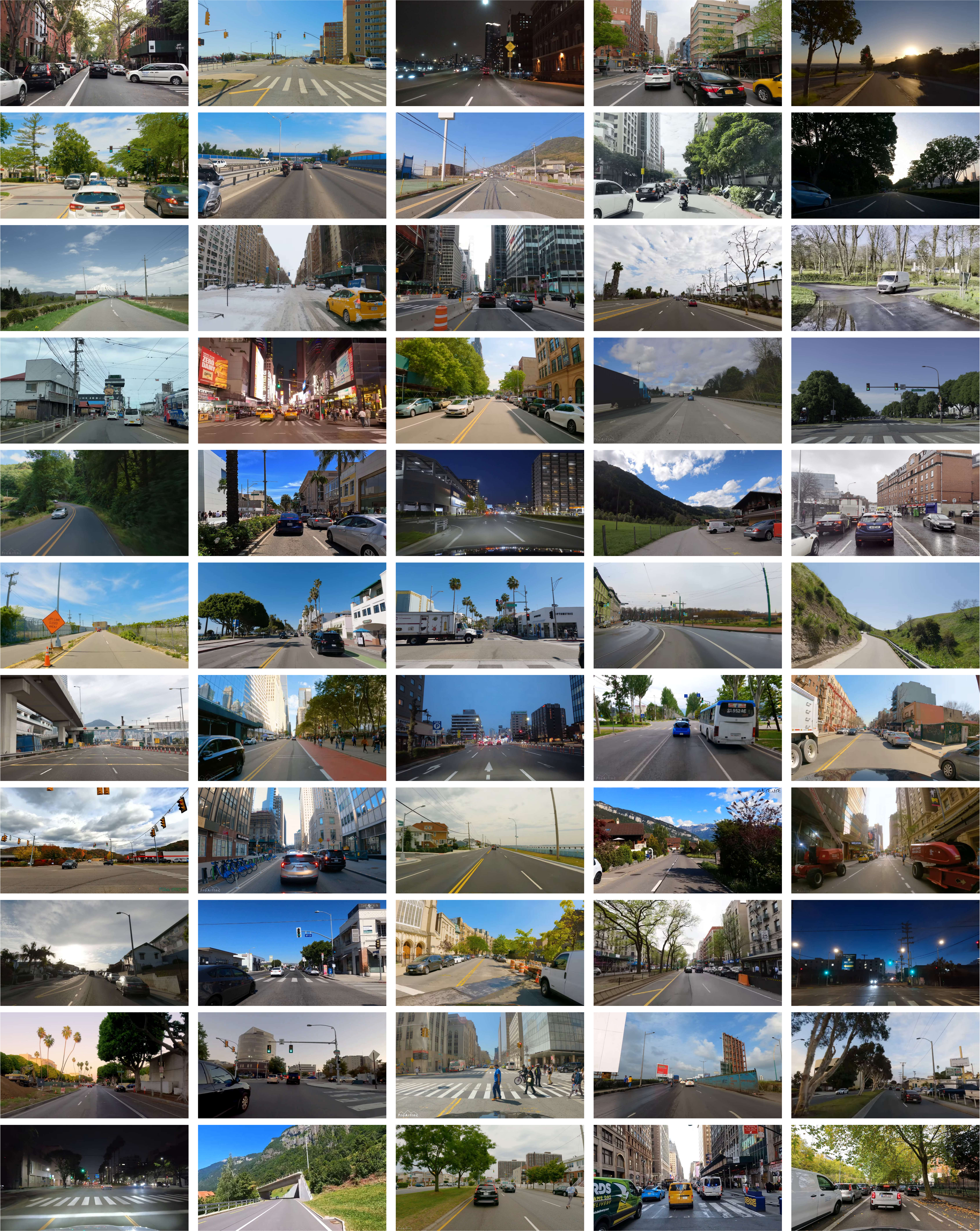}
\caption{
\textbf{Showcase on the Diversity of Open-world Data for Pre-training.}
The images we used for pre-training come from over 700 cities and 3000 hours of data. 
The vast amount and diversity of data ensure the model's generalizability.
}
\label{fig:raw_img}
\end{figure}

\begin{figure}[H]
\centering
\vspace{1.5cm}
\includegraphics[width=0.99\linewidth]{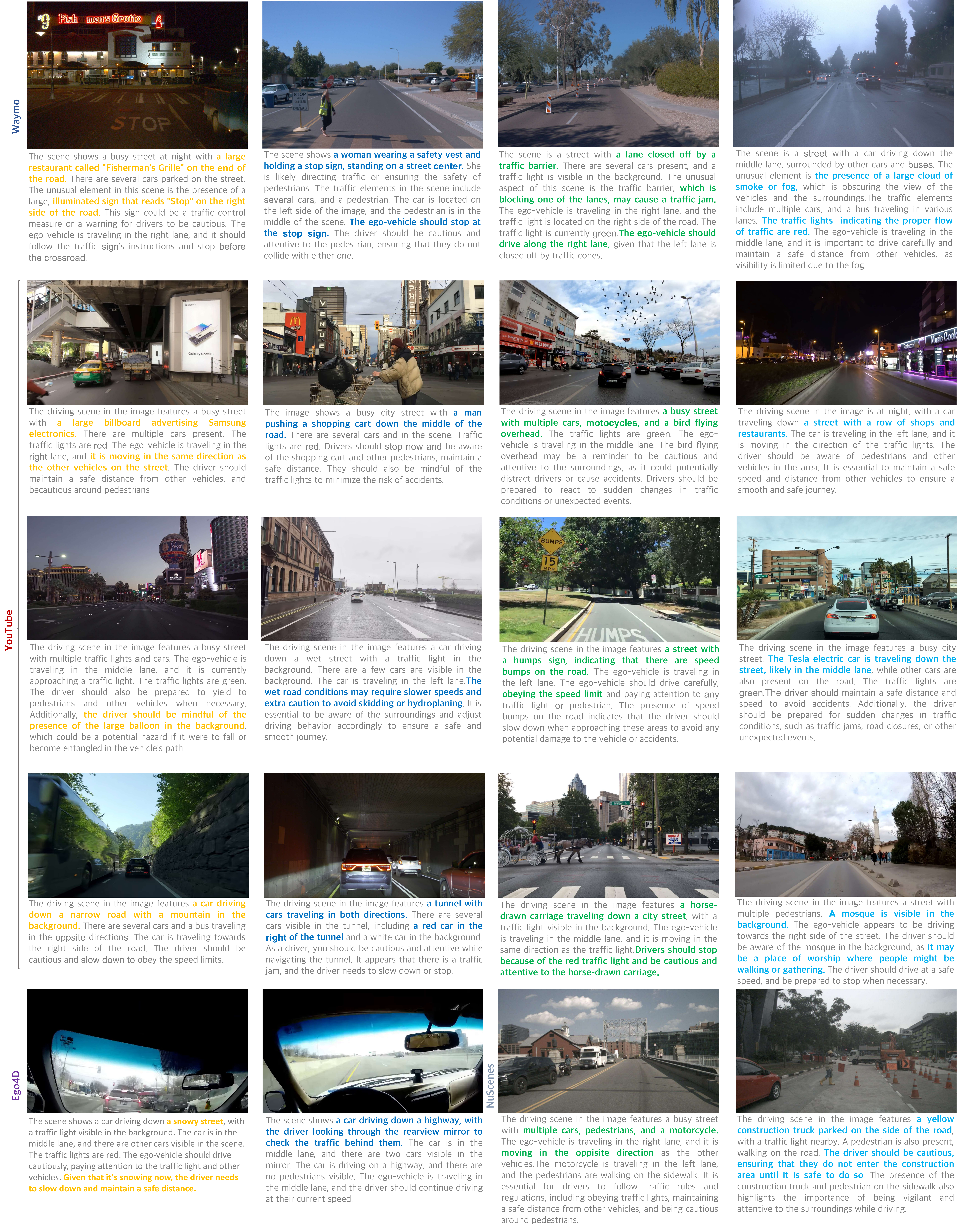}
\caption{
\textbf{The visualization of description annotation.}
These labels are generated through LLaMA-Adapter V2 \cite{gao2023llamaadapterv2}, followed by human quality check and revision.
}
\label{fig:description_label}
\end{figure}

\smallskip \noindent \textbf{Demonstration of data diversity.}
The diversity of data is a prerequisite to ensure that the model has a good generalization ability. 
In \cref{fig:raw_img}, we display the richness of the pre-training images from over seven hundred cities, covering various lighting, environment, weather, road conditions, \textit{etc}. 
\cref{fig:description_label} shows the diverse description labels we generated, guaranteeing the model's reliability in describing new scenarios.

\subsection{Embodied Scene Understanding}

\begin{figure}[H]
\centering
\vspace{1.5cm}
\includegraphics[width=0.99\linewidth]{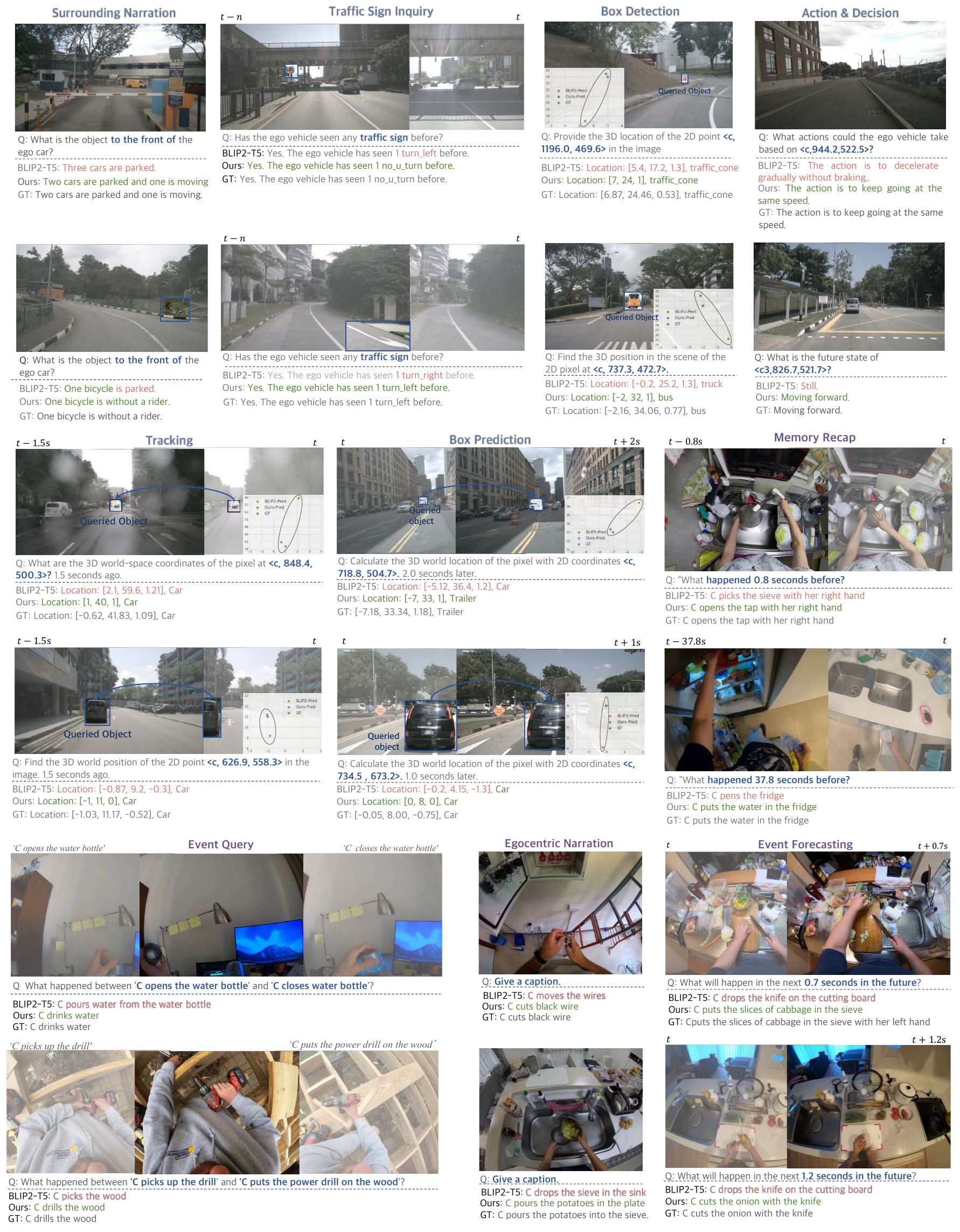}
\caption{
\textbf{Visualization on embodied understanding benchmark.}
}
\label{fig:ten_tasks}
\end{figure}

We offer more visualizations of ten tasks in \cref{fig:ten_tasks} to prove ELM's superiority on the benchmark further.
This figure indicates that ELM has significantly improved the reasonableness and precision of responses compared to the BLIP2-flant5 \cite{li2023blip2} baseline.

\subsection{Planning}

\begin{figure}[H]
\centering
\vspace{2.2cm}
\includegraphics[width=0.99\linewidth]{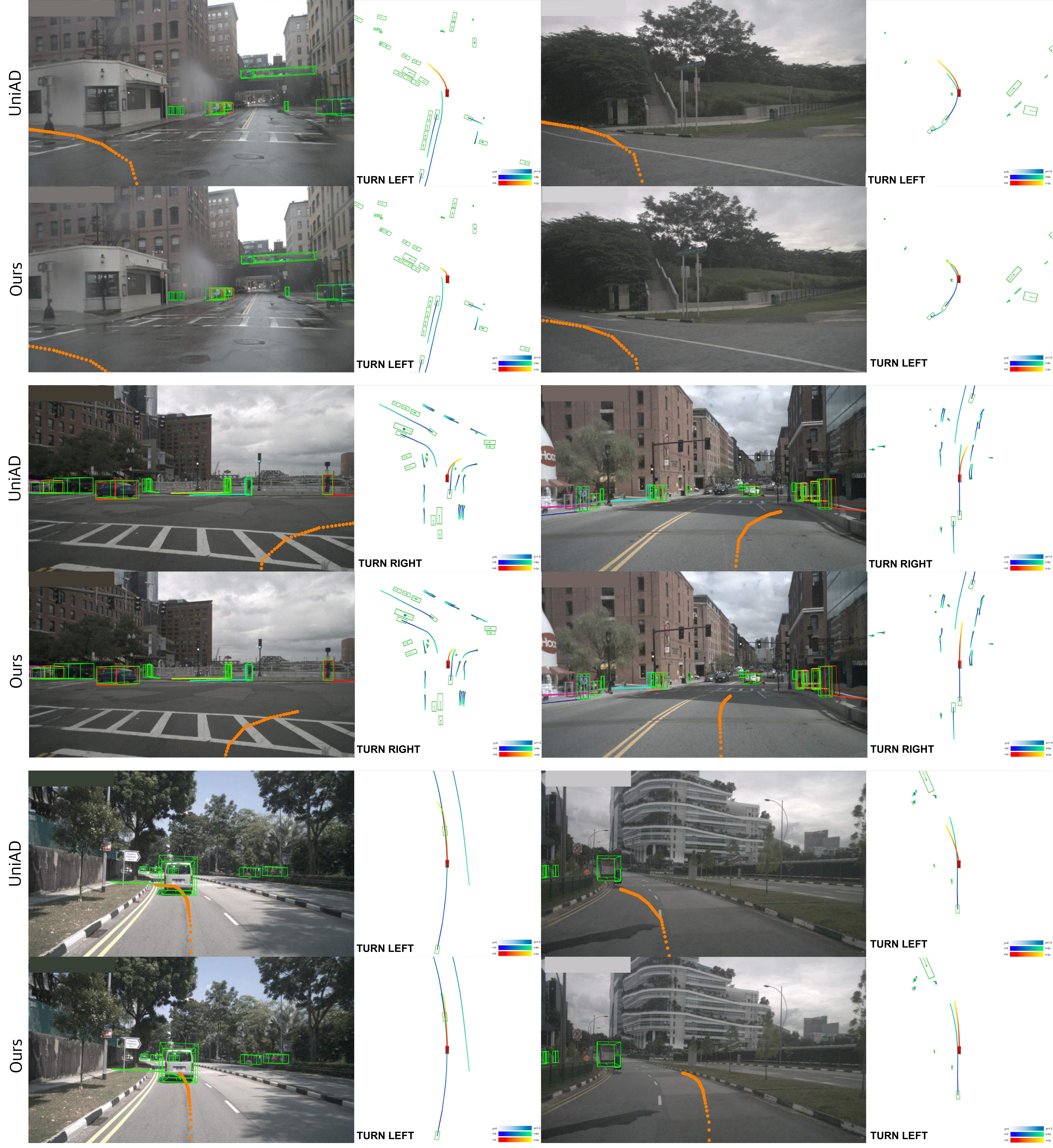}
\caption{
\textbf{Visualization on planning.}
We compared the trajectory prediction results of UniAD \cite{hu2023_uniad} and ELM in some relatively complex scenarios. The results show that ELM has the potential to follow traffic rules and maintain the current lane.
}
\label{fig:planning}
\end{figure}

In \cref{fig:planning}, we visualize ELM's planning results with images and instruction inputs.
In the first row of comparisons, ELM avoids driving towards the roadside and into opposing traffic lanes during turns.
It can be seen that ELM has superior abilities to comply with traffic rules compared to UniAD \cite{hu2023_uniad}. 
In the third group of comparisons, when ELM receives a \texttt{turn\_right} instruction, it considers the intersection scenario and opts for slow driving. 
Experiments from the fourth to the sixth groups demonstrate that ELM maintains its current lane while following instructions. 
These comparison experiments illustrate that ELM integrates common sense and shows the potential to outperform UniAD in complex scenarios.

\subsection{Multi-round Dialogues}

\begin{figure}[H]
\centering
\vspace{1.2cm}
\includegraphics[width=0.99\linewidth]{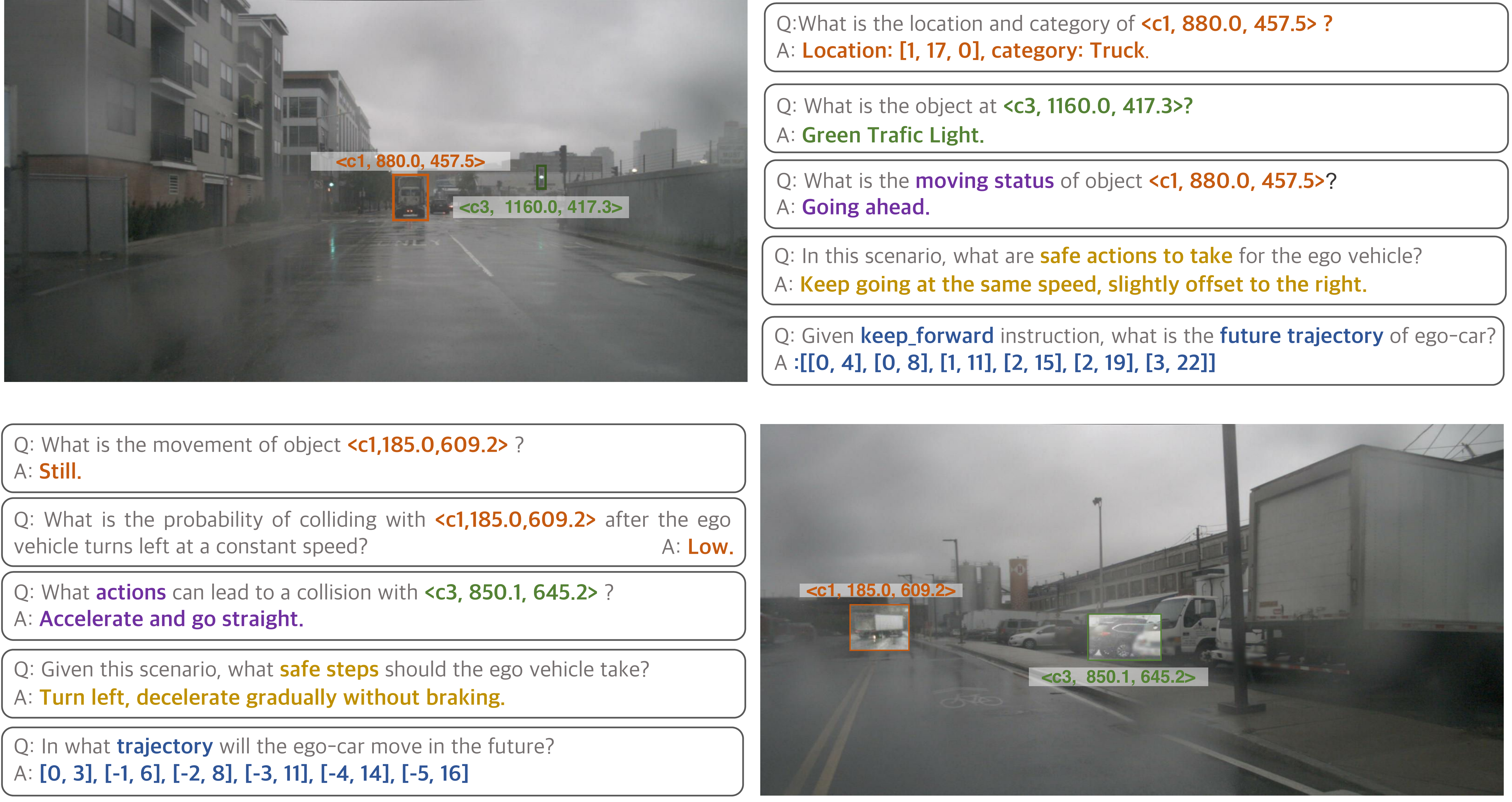} % No compression needed
\caption{
\textbf{Visualization on multi-round dialogues.}
We provide application cases of ELM in real scenarios, where, through multiple rounds of dialogue, it goes from describing driving scenarios to generating high-level control instructions and  
% producing 
trajectories at the end.
}
\vspace{-0.8cm}
\label{fig:multi_round}
\end{figure}

As an example of ELM's application in a practical scenario, we provide in \cref{fig:multi_round} the process of ELM having multiple rounds of dialogue with the user, gradually understanding the scenario, and ultimately making a decision.
This round of dialogue begins with identifying traffic elements in the scene, including vehicles, pedestrians, and traffic signs, followed by inquiries about the movement of the objects and extrapolates reasonable higher-order driving instructions.
Ultimately, the model can provide a planning result, \textit{i.e.}, future trajectory points.

% \subsection{Zero-shot on New Scenarios}

% \cref{fig:zero_shot} shows the visualization results of ELM performing Box Detection and Planning in new scenarios.
% %
% It can be observed that, even in cases the model has never encountered, it can still produce reasonable results.
% %
% In such situations, traditional detection models need help to perform well.
% %
% The results indicate that ELM, with its common sense reasoning ability, has the potential to outperform traditional methods in unseen scenarios and unpredictable situations.

\section{Licensing and Privacy Considerations}

All the data is under the CC BY-NC-SA 4.0 license\footnote{https://creativecommons.org/licenses/by-nc-sa/4.0/deed.en}.
Other datasets (including nuScenes \cite{qian2023nuscenes} and Ego4D \cite{grauman2022ego4d}) inherit their own distribution licenses.

We place a high value on license and privacy protection, following the precedent from YouTube-8M \cite{abu2016youtube}, YouTube-VOS \cite{xu2018youtube}, AOC \cite{zhang2022learning}, CelebV-HQ \cite{zhu2022celebvhq}, and Kinetics \cite{kay2017kinetics}, \textit{etc}. 
For videos from YouTube, permission to the video content is received through a Creative Commons license.
Besides, we skip channel-related content at the beginning and end of the videos during data processing to ensure we do not infringe upon the rights of logos, channel owner information, or other copyrighted materials.
We do not provide video content; users are redirected to original YouTube videos via a link. 
The platform safeguards personal info with encryption, access limits, and identity checks to prevent unauthorized video access. 
We will credit the source, provide a link to the license, and state that no modifications have been made to the video itself but only the labeled text, and that the data will not be used for commercial purposes.
All the data we obtain complies with regulations and YouTube's Privacy Policy.

In addition, we comply with any limitations required by applicable law and any requests submitted by users. For instance, users may have the right to view, correct, and delete personal information we possess about them, such as deleting text labels, unlinking videos, and de-identifying data.
For further protecting personally identifiable information, we use off-the-shelf tools to add mosaics to faces and license plates in the videos during the data post-processing similar to some datasets \cite{wang2019apolloscape,grauman2022ego4d}.
This measure can effectively prevent the model from memorizing identifiable information during the training process.
Based on existing standards \cite{grauman2022ego4d}, we take all possible measures to ensure data privacy.

\section{Discussions}

\smallskip \noindent \textbf{Limitations.}
Our ELM focuses on understanding driving scenes and interacting with human users, leaving much room for future exploration and expansion in other areas. 
A significant limitation is that the current model does not generate physical control signals as referred to \cite{chen2023drivingwithllms}. 
Our model cannot implement the behaviors it generates, unlike robotic models \cite{rt22023arxiv,shah2023vint,brohan2022rt,gu2023robotic} that can directly manipulate their environment.
%
% Such direct interaction with the environment is an essential aspect of embodied cognition, which is a fundamental principle in both human cognition and robotic operation.
%
% In robot-related fields, models not only generate actions but also execute them, modifying the state of their surroundings and effectively learning from the resulting changes. 
% By contrast, our model currently misses real-time feedback from the environment. 
Direct interaction provides a valuable channel for the model to optimize its behaviors based on environmental responses. Thus, the absence of this component might limit the model's potential to learn and adapt effectively.
Therefore, introducing the ability for the model to interact with its environment directly could be an essential step in future work, potentially leading to significant improvements in the model's performance and adaptability.

While the prospect of implementing a prototype system to transform our ELM into an embodied agent for closed-loop autonomous driving sounds promising, it is worth reiterating that this plan is quite forward-looking.
Integrating the system into an autonomous driving framework entails overcoming technical challenges related to hardware compatibility and real-time responsiveness.
Additionally, the potential for unforeseen practical obstacles or the emergence of unpredictable patterns during driving should be considered. 
While ELM, in its current form, can theorize and understand different driving scenarios, its performance in a more dynamic, real-world environment still needs to be tested.

The effectiveness of common sense reasoning aiding decision-making in novel scenarios still needs more adequate validation. 
While it is theorized that common sense reasoning can provide valuable insights during unpredictable or unusual situations, we have yet to build an efficient metric to measure the impact and benefit accurately.
These are all important considerations that constitute limitations in our current model and provide directions for our future work.

\smallskip \noindent \textbf{Risks in automated labeling.}
Apart from human-annotated labels, ELM also uses extensive automated labeling and templated data to provide ground truth labels, which may raise concerns about substantial risk in using automated labels to train new automated systems.
Encouragingly, the success of DALLE-3 \cite{openai2023dalle3} showcases the widely introducing auto labeling can boost model performance, given manual labeling diversity is constrained.
It asserts that training with a mix of real and automatically annotated data can effectively reduce data bias caused by limited manual annotating (such as personal grammar habits and idiosyncrasies), given the creativity and text diversity of large language models.
%
% Following [{\color{cvprblue}1}], ELM achieves superior quality assurance to others [{\color{cvprblue}8}, {\color{cvprblue}12}, {\color{cvprblue}50}, {\color{cvprblue}52}] through human filtering and verification.
% 
% Recent study \cite{} verifies that human filtering and verification are effective approaches to improving the quality of automated labeling data.
% With these, we achieve a quality assurance strategy that is theoretically superior to others [{\color{cvprblue}8}, {\color{cvprblue}12}, {\color{cvprblue}50}, {\color{cvprblue}52}].
%
Besides, a recent study \cite{chung2023increasing} verifies that human filtering and verification are effective approaches to improving the quality of automated labeling data.
Results in Tab. {\color{red} 5} of main paper indicate ELM achieves theoretically optimal quality assurance via human-AI cooperation, aligning with their conclusion.
Future work will examine logit suppression, temperature sampling, \textit{etc}, to reduce automated system risks.

\smallskip \noindent \textbf{Issues of statistical shortcuts.}
It is known that deep learning models can resort to statistical shortcuts in such tasks, potentially guessing answers based on the majority of training samples rather than truly understanding the context in the provided video. 
This may raise concerns about do today’s VLMs understand and reason about the visual world as well as LLMs understand text-based worlds.
Thankfully, the Egocentric Narration task involves changing images and fixed text as input. 
Thus, its result (\textbf{26.5}) in Tab. {\color{red}3} of the main paper already affirms ELM's visual comprehension.
Further experiments verify ELM does not guess answers from training samples, as replacing images with fixed ones causes the results to drop by more than half.

\begin{figure}[H]
\centering
\includegraphics[width=0.99\linewidth]{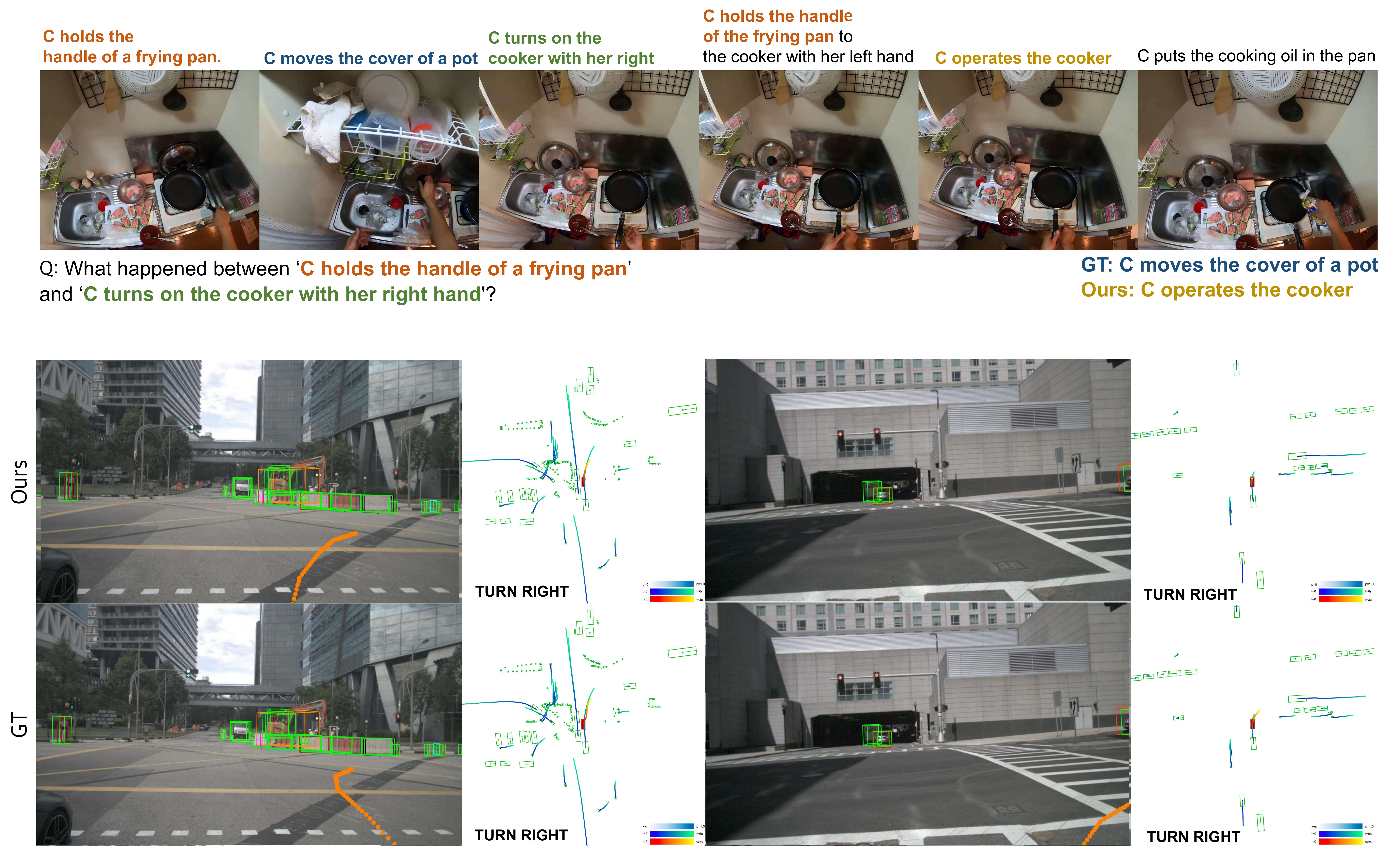}
\caption{
\textbf{Failure cases demonstration on Event Query and Planning tasks.}
}
\label{fig:bad_case}
\end{figure}

\smallskip \noindent \textbf{Failure cases.}
\cref{fig:bad_case} shows several cases where our model fails.
The upper part of the figure presents cases where ELM performs poorly in the Event Query task.
Particularly, the model's discriminative capacity declines when similar behaviors appear in the scene (as in the first and fourth images), thereby failing to locate the specific event referred to in the query. 
Despite the different image content represented in the third and sixth images, the capacity of FlanT5 \cite{2020t5} limits ELM's recognition of the difference between them.
The lower part of the figure shows two cases where ELM's planning does not align with the ground truth. 
The image on the left indicates that ELM tends to opt for a more conservative planning strategy, proactively starting to evade when vehicles and pedestrians appear on the road to ensure driving safety. 
The image on the right shows that ELM possesses a latent ability to understand common sense, going beyond obeying directives. 
Upon recognizing the red light, the model chooses not to execute the command to turn right, opt to stop and wait instead.

\newpage

% \newpage

% % {
% %     \small
% %     \bibliographystyle{ieeenat_fullname}
% %     \bibliography{short,main}
% \bibliography{main}

% \end{document}

\end{document}